\documentclass[11pt]{article}
\usepackage[utf8]{inputenc}

\usepackage[T1]{fontenc}    
\usepackage{url}            
\usepackage{booktabs}       
\usepackage{amsfonts}       
\usepackage{nicefrac}       
\usepackage{microtype}      
\usepackage{amsmath,amssymb}
\usepackage{graphicx}
\usepackage{subcaption}
\usepackage{tabularx}
\usepackage{array}
\usepackage[font=scriptsize]{caption}
\usepackage{enumerate}
\usepackage[OT1]{fontenc}
\usepackage{natbib}
\usepackage[usenames]{color}
\usepackage[colorlinks,
linkcolor=red,
anchorcolor=blue,
citecolor=blue
]{hyperref}
\usepackage{mathrsfs}
\usepackage{fullpage}
\usepackage{hyperref}

\let\oldabstract\abstract
\let\oldendabstract\endabstract
\makeatletter
\renewenvironment{abstract}
{
\oldabstract}
{\oldendabstract}
\makeatother

\begin{document}

\title {\huge The Bayesian Method of Tensor Networks}

\author
{Erdong Guo\thanks{University of California, Santa Cruz, email: \texttt{eguo1@ucsc.edu}.} \qquad David Draper\thanks{University of California, Santa Cruz, email: \texttt{draper@ucsc.edu}} 
}

\date{}

\maketitle

\begin{abstract}
Bayesian learning is a powerful learning framework which combines the external information of the data (background information) with the internal information (training data) in a logically consistent way in inference and prediction. By Bayes rule, the external information (prior distribution) and the internal information (training data likelihood) are combined coherently, and the posterior distribution and the posterior predictive (marginal) distribution obtained by Bayes rule summarize the total information needed in the inference and prediction, respectively. 
In this paper, we study the Bayesian framework of the Tensor Network from two perspective. 
First, we introduce the prior distribution to the weights in the Tensor Network and predict the labels of the new observations by the posterior predictive (marginal) distribution. Since the intractability of the parameter integral in the normalization constant computation, we approximate the posterior predictive distribution by Laplace approximation and obtain the out-product approximation of the hessian matrix of the posterior distribution of the Tensor Network model. 
Second, to estimate the parameters of the stationary mode, we propose a stable initialization trick to accelerate the inference process by which the Tensor Network can converge to the stationary path more efficiently and stably with gradient descent method. 
We verify our work on the MNIST, Phishing Website and Breast Cancer data set. We study the Bayesian properties of the Bayesian Tensor Network by visualizing the parameters of the model and the decision boundaries in the two dimensional synthetic data set.
For a application purpose, our work can reduce the overfitting and improve the performance of normal Tensor Network model.
\end{abstract}

\section{introduction}
Bayesian Learning is a framework to combine the internal data information with your background knowledge on your specific learning task, namely the external information in a logically consistent way by Bayes Theorem \cite{draper2006bayesian, draper2013bayesian}. 
In the Bayesian framework, the parameters of the model are not fixed constants but random variables whose distributions should be introduced according to the background information. 
From a information perspective, the background information (the knowledge) external to the data set is 'injected' into the learning model optimally by Bayes Theorem \cite{zellner1988optimal}. 

In this framework, parameters are not estimated by point estimator such as the Maximum Likelihood Estimator (M.L.E.) instead the posterior distribution of the parameters are given. 
In this step, the technical difficulty is the computation of the normalization constant. 
In some application cases, people use the Dirac distribution of the optimal mode to approximate the posterior distribution (M.A.P.) which is equivalent to use the Maximum A Posterior Estimator to do point estimation of the parameters in the model \cite{nowlan1992simplifying}. 
To get better approximation of the posterior distribution, higher order term should be remained and till second order terms are preserved and Normal distribution is obtained by Laplace approximation trick. 



In our work, we proposed a robust initialization method to infer the parameters of the Bayesian Tensor Network model more quickly. Since the special structure of the Tensor Network which is a sum of tensor chain products, the output of this model is easily blowing out or decaying to zero which will lead to the instability of the inference process. Using our initialization strategy, the parameters and their gradients will stay in the stable region where gradient descent optimization will behave healthily. 
To predict the new observations, we get the predictive marginal posterior distribution of Bayesian Tensor Network model by Laplace Approximation trick, namely second order approximation. We observe that the hessian matrix of the Tensor Network model has a nice analytical expression which will be computed more efficiently than the Bayesian Neural Network. Hopefully, the Bayesian Tensor Network can be easily scaled to much larger enterprise level applications.

In practical application, labels will be assigned to the new observations in classification problem.
To determine the decision boundary, the utility matrix (negative loss matrix) will be written down and then the decision threshold will be determined by minimize the expected loss.



\section{Related Work}
Deep Neural Networks have made great achievements in recent years \cite{lecun2015deep, hinton2006fast, lecun1995convolutional, krizhevsky2012imagenet, hochreiter1997long, srivastava2014dropout}. Actually several work were proposed to develop the Bayesian Framework of Neural Network. In the Bayesian Framework of Neural Network, the key part and also difficulty part is to approximate the posterior distribution and also the predictive marginal distribution.

In the framework of hierarchical modeling for model uncertainty which is published by another author of this paper, Prof. David Draper proposed a general framework to consider the model uncertainty and also discussed the computation techniques in inference and prediction steps in the Bayesian Modeling considering model uncertainty \cite{draper1995assessment}. 
At the same time, some other early pioneers of Bayesian Neural Networks also contributes great to this area. David MacKay focus on the Bayesian Neural Networks as an adaptive learning model which means he used the Bayesian Framework to do Neural Networks model comparing and the hyperparameters and model structures selection by the evidence framework \cite{mackay1992interpolation}. Radford Neal focus on constructing appropriate priors for the Neural Networks based on the properties of the Neural Networks. Neal suggested to use the infinite hidden units network and the priors converge to the stochastic process. Neal also proposed to use the Hybrid Monte Carlo to do the posterior predictive distribution (integral) simulation \cite{neal2012bayesian}. Instead of approximate the posterior integral, variational method idea, namely searching in the parameterized functional space by maximizing the Evidence Lower Bound is widely explored \cite{blei2006variational, kingma2013auto}.
Recently, there are amounts of developments following and developing above ideas \cite{buntine1991bayesian, tran2019bayesian, lee2017deep, arora2019exact, arora2019harnessing, blundell2015weight, xiong2011bayesian, salakhutdinov2008bayesian, balan2015bayesian, hernandez2015probabilistic}. 

Tensor Networks are original proposed to describe the quantum many-body states \cite{orus2019tensor, chabuda2020tensor, glasser2020probabilistic}. 
In the last several years, several work were proposed such as using the Matrix Product States (MPS) to construct new Learning Models \cite{novikov2015tensorizing, stoudenmire2016supervised, han2018unsupervised}. 

\section{Initialization of Tensor Network}
\label{initialization_tnn}
\subsection{Background and Set Up}
Although Tensor Network has powerful representation ability because of the bond dimension between every pair of tensor nodes neighboring each other, it is difficult to train long Matrix Product States chains. 
Since every tensor node in the chain will contribute one multiplication factor, the output will roughly increase or decrease with a exponential rate with respect to the number of nodes. 

The Set up of the Tensor Network is as follows,
\begin{align}
    \psi^{l}(\mathbf{x}) = \sum_{\{\alpha_{i}, s_{i}\}}A^{s_{1}}_{\alpha_{1}\alpha_{2}}\cdots A^{s_{i}l}_{\alpha_{i}\alpha_{i+1}}\cdots A^{s_{n}}_{\alpha_{n}\alpha_{1}}\Phi^{s_{1}\cdots s_{n}}(\mathbf{x}),
\end{align}
where
\begin{align*}
    \Phi^{s_{1}\cdots s_{n}}(\mathbf{x}) = \phi^{s_{1}}(x_{1})\cdots\phi^{s_{i}}(x_{i})\cdots\phi^{n}(x_{n}).
\end{align*}
We can get a rough estimation of the amplitude of $\psi^{l}(\mathbf{x})$ as 
\begin{align}
    &|\psi^{l}( \mathbf{x})| \sim s^{n}\alpha^{n}O(C^{n}), 
\end{align}
where 
\begin{align*}
\quad s = |\{s_{i}\}|, \quad \alpha = |\{\alpha_{i}\}|.
\end{align*}
With the assumption that $|A^{s_{i}}_{\alpha_{i}\alpha_{i+1}}| = O(C)$, and $|\phi^{s_{i}}(x_{i})| \sim O(1)$,
where $\alpha$ represents the bond dimension, $s$ represents the dimension of the kernel space and $n$ represents the number of the nodes.

The networks will not work well or even not work if the tensor nodes are not well initialized since little perturbation of the tensor weights will be amplified because of the exponential rate respect to the number of nodes. 

To solve the respond or gradients exploding or vanishing problem, we focused on initialize the parameters with the appropriate distribution. 
To train deep neural networks, Glorot and Bengio \cite{glorot2010understanding} proposed to initialize the parameters with the scaled uniform distribution in the Sigmoid neuron case. 
For the Relu neuron, He and his collaborators \cite{he2015delving} proposed to use the Gaussian distribution with neural number determined variance. 
In the following section, we will derive the variance of the Gaussian distribution used in our initialization method. 

\subsection{Variance Analysis}
We use the Gaussian distribution to initialize Bayesian Tensor Networks model and set the mean to be zero since we want the respond to stay around zero. For the variance, we need to analyze the variance of the respond as the function of the variance of the tensor weights. 

\begin{align}
\mathrm{Var}[\psi^{l}(\mathbf{x})] 
&= \mathrm{Var}[\sum_{\{s, \alpha\}}A^{s_{1}}_{\alpha_{1}\alpha_{2}}\cdots A^{s_{i}l}_{\alpha_{i}\alpha_{i+1}}\cdots A^{s_{n}}_{\alpha_{n}\alpha_{1}}\Phi^{s_{1}\cdots s_{n}}(\mathbf{x})]\\
&=\sum_{\{s\}}\alpha^{n}\mathrm{Var}[A^{s}]^{n}\mathrm{Var}[\phi^{s}(x)]^{n}
\end{align}

In above derivation, we assume that all the components of the tensor weights are independent and identically distributed (I.I.D.) random variables. 
For the kernel part, we assume that all the components of input kernel are also I.I.D.. 

Considering the particular kernel function $\phi(x) = [x, 1-x]^{T}$, we have
\begin{align}
&\mathrm{Var}[\psi^{l}(\mathbf{x})] = s^{n}\alpha^{n}\mathrm{Var}[A]^{n}\mathrm{Var}[x]^{n}.
\end{align}
From the gradients perspective, we can write down the gradients of tensor node $A_{\alpha_{i},\alpha_{i+1}}^{s_{i}}$ as follows.
\begin{align}
[\nabla\psi^{l}(\mathbf{x})]^{l}_{\alpha_{i}\alpha_{i+1}} = \sum_{\{s, \alpha\}\backslash\{s_{i}, \alpha_{i}, \alpha_{i+1}\}}A^{s_{1}}_{\alpha_{1}\alpha_{2}}\cdots A^{s_{i-1}}_{\alpha_{i-1}\alpha_{i}}A^{s_{i+1}}_{\alpha_{i+1}\alpha_{i+2}}\cdots A^{s_{n}}_{\alpha_{n-1}\alpha_{n}}\Phi^{s_{1}\cdots s_{i-1}s_{i+1}\cdots s_{n}}(\mathbf{x}).
\end{align}
We get the variance of the gradients as 
\begin{align}
\mathrm{Var}[\nabla\psi^{l}(\mathbf{x})] = s^{n-1}\alpha^{n-2}\mathrm{Var}[A]^{n-1}\mathrm{Var}[x]^{n}.
\end{align}

In a good initialization method, we can set the variance of the tensor nodes $\mathrm{Var}[A^{s_{i}}_{\alpha_{i}\alpha_{j}}]$ as the same order of the output variance $\mathrm{Var}[\psi^{l}(\mathbf{x})]$ which will avoid increasing or decreasing the response exponentially, so we get
\begin{align}
\mathrm{Var}[\psi^{l}(\mathbf{x})]\sim\mathrm{Var}[A]\to \mathrm{Var}[A] \sim O((s\alpha)^{-\frac{n}{n-1}}\mathrm{Var}[x]^{-\frac{n}{n-1}}) 
\end{align}
From the gradient perspective, the variance of the gradients $\mathrm{Var}[\nabla\psi^{l}(\mathbf{x})]$ is a same order number as the variance of the tensor weights $\mathrm{Var}[A^{s_{i}}_{\alpha_{i}\alpha_{j}}]$, then we obtain
\begin{align}
\mathrm{Var}[\nabla\psi^{l}(\mathbf{x})]\sim \mathrm{Var}[A]O(C)\to \mathrm{Var}[A] \sim O((C^{\frac{1}{n-2}}s^{-\frac{n-1}{n-2}}\alpha^{-1}\mathrm{Var}[x]^{-\frac{n}{n-2}})     
\end{align}

We analyze the asymptotic behavior of our initialization formula. From the tensor weights perspective,
\begin{align}
    \mathrm{Var}[A]_{n\to\infty} = O((s\alpha\mathrm{Var}[x])^{-1}).
\end{align}
From the gradient perspective,
\begin{align}
\label{eq: var_initialization_formula}
   \mathrm{Var}[A]_{n\to\infty} = O((s\alpha\mathrm{Var}[x])^{-1}.
\end{align}

In the large $n$ limit, we get the same initialization variance of the tensor weights from weights perspective and gradient perspective which means our initialization method is logically consistent. 

Different from Xavier \cite{glorot2010understanding}, and He initialization \cite{he2015delving}, our initialization formula does not depend on the number of the nodes heavily as we see the number of nodes $n$ just get into the index of the variance of each feature of the data as a fractional factor and the factor will converge to $1$ quickly. However, the bond dimension and the physical dimension, namely the dimension of the kernel has a huge effects on the initialization.

We analyse the mean of the parameters following the above idea,

\begin{align}
\mathrm{E}[\psi^{l}(\mathbf{x})] &= \mathrm{E}[\sum_{\{\alpha_{i}, s_{i}\}}A^{s_{1}}_{\alpha_{1}\alpha_{2}}\cdots A^{s_{i}l}_{\alpha_{i}\alpha_{i+1}}\cdots A^{s_{n}}_{\alpha_{n}\alpha_{1}}\Phi^{s_{1}\cdots s_{n}}(\mathbf{x})]\\
&=s^{n}\alpha^{n}\mathrm{E}[A^{s_{i}}_{\alpha_{i-1}\alpha_{i}}]^{n}\mathrm{E}[\phi(x_{i})]^{n},
\end{align}

In most practical case, the training data set is usually pre-processed and the mean of the training data set is usually transferred to $0$.
So we have 
\begin{align}
    \mathrm{E}[\phi(x_{i})] = 0,
\end{align}

then we get 
\begin{align}
\mathrm{E}[\psi^{l}(x)] = 0.    
\end{align}

Similarly, we can get 
\begin{align}
    \mathrm{E}[\nabla\psi^{l}(x)] = 0. 
\end{align}

This means that the mean of the responses and the gradients will always stay in healthy region, namely $0$ no matter what the initialization distribution is. So we do not need to care about the mean in the initialization method design.

\subsection{Numerical Results}
In this section. we study the performance of our formula on the MNIST \cite{lecun1998mnist}, Phishing Website and Breast Cancer data set \cite{Dua:2019} and also we compare our initialization method with Xavier and He initialization methods.

\begin{figure}[!ht]
\begin{subfigure}{.49\textwidth}
  \centering
  \includegraphics[width=.9\linewidth]{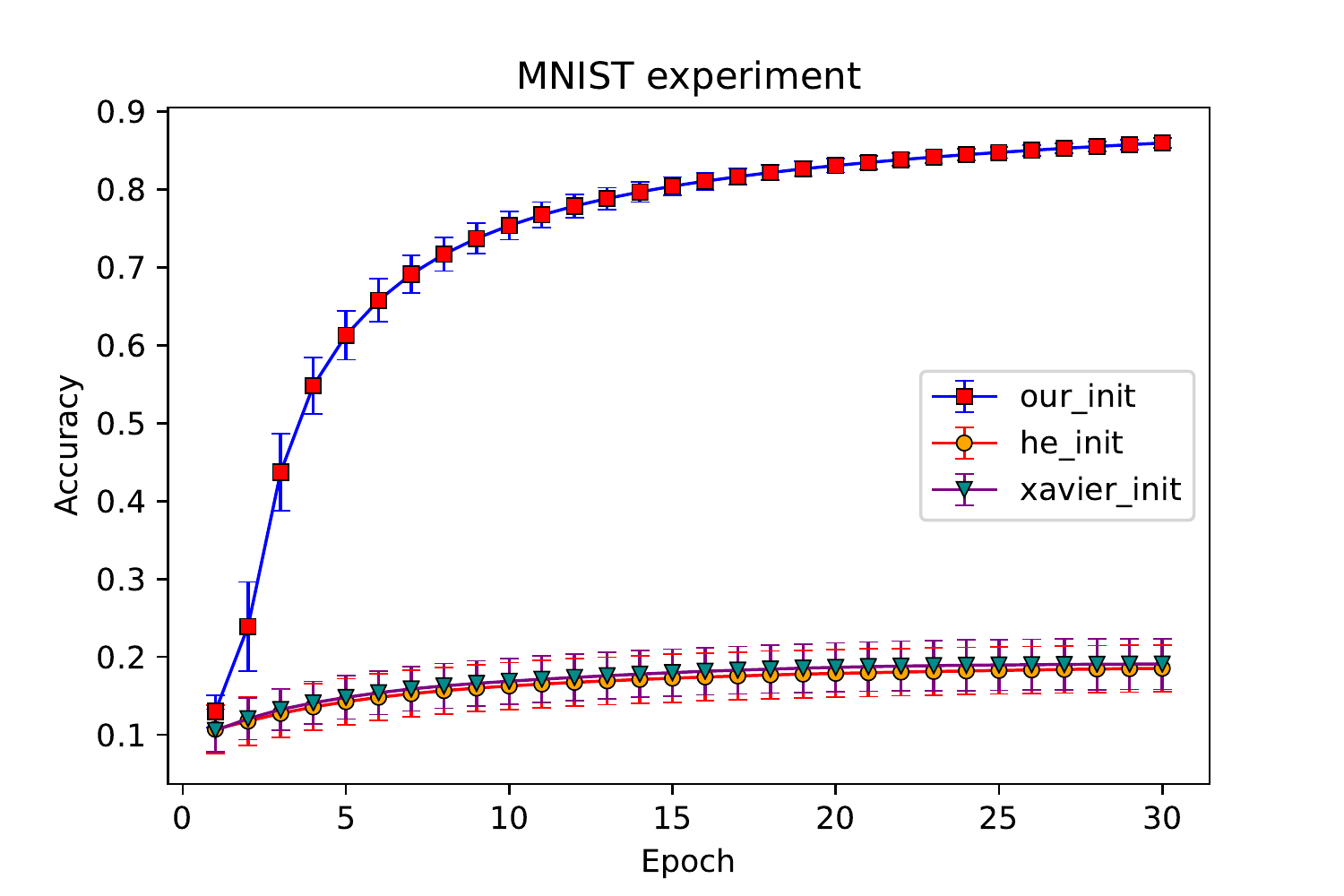}
  \caption{Initialization method comparing on MNIST Data Set}
  \label{fig: mnist_init_comparing}
\end{subfigure}
\begin{subfigure}{.49\textwidth}
  \centering
  \includegraphics[width=.9\linewidth]{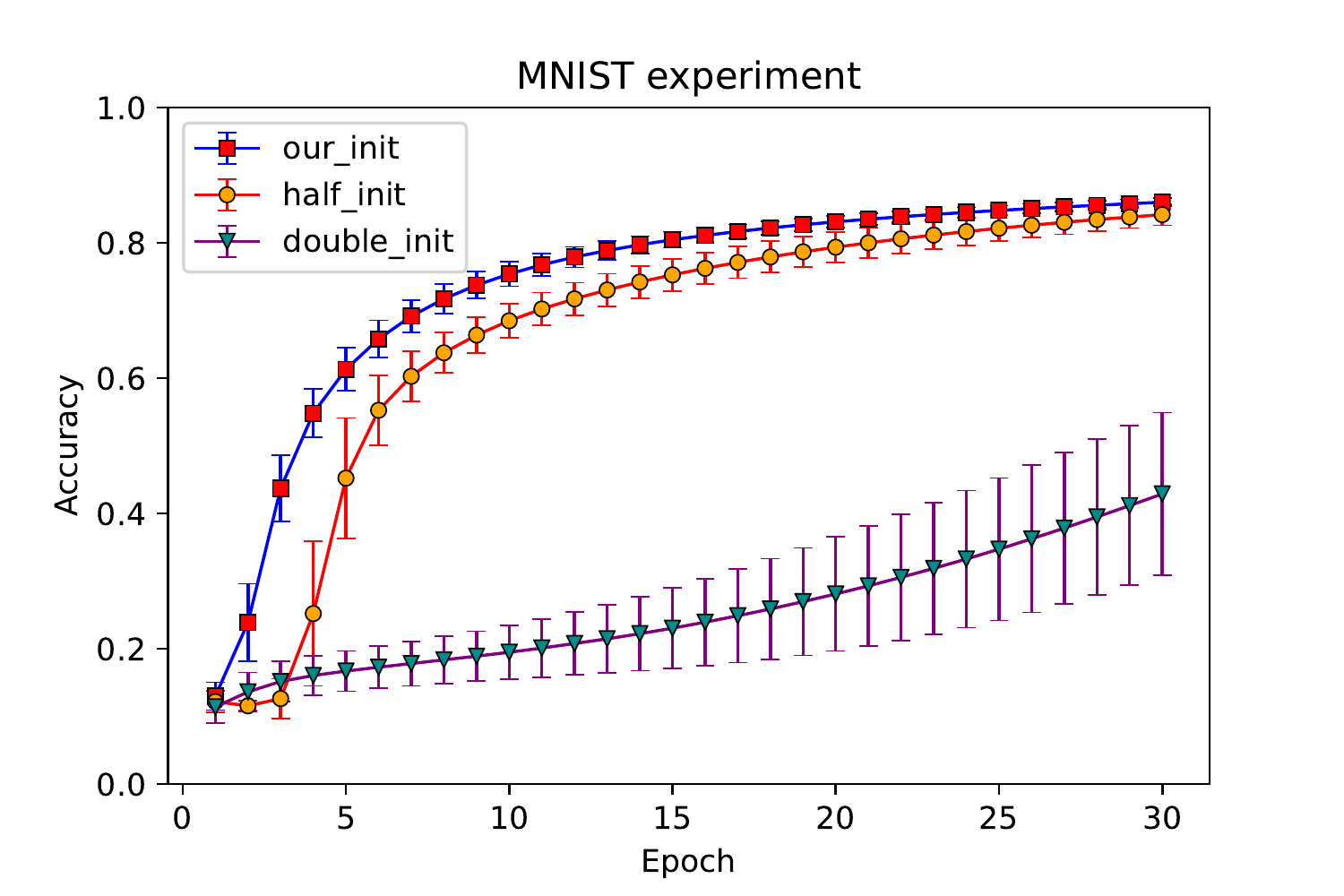}
  \caption{Initialization with $\sigma$, $\frac{1}{2}\sigma$ or $2\sigma$}
  \label{fig: mnist_init_std_comparing}
\end{subfigure}\\
\begin{subfigure}{.49\textwidth}
  \centering
  \includegraphics[width=.9\linewidth]{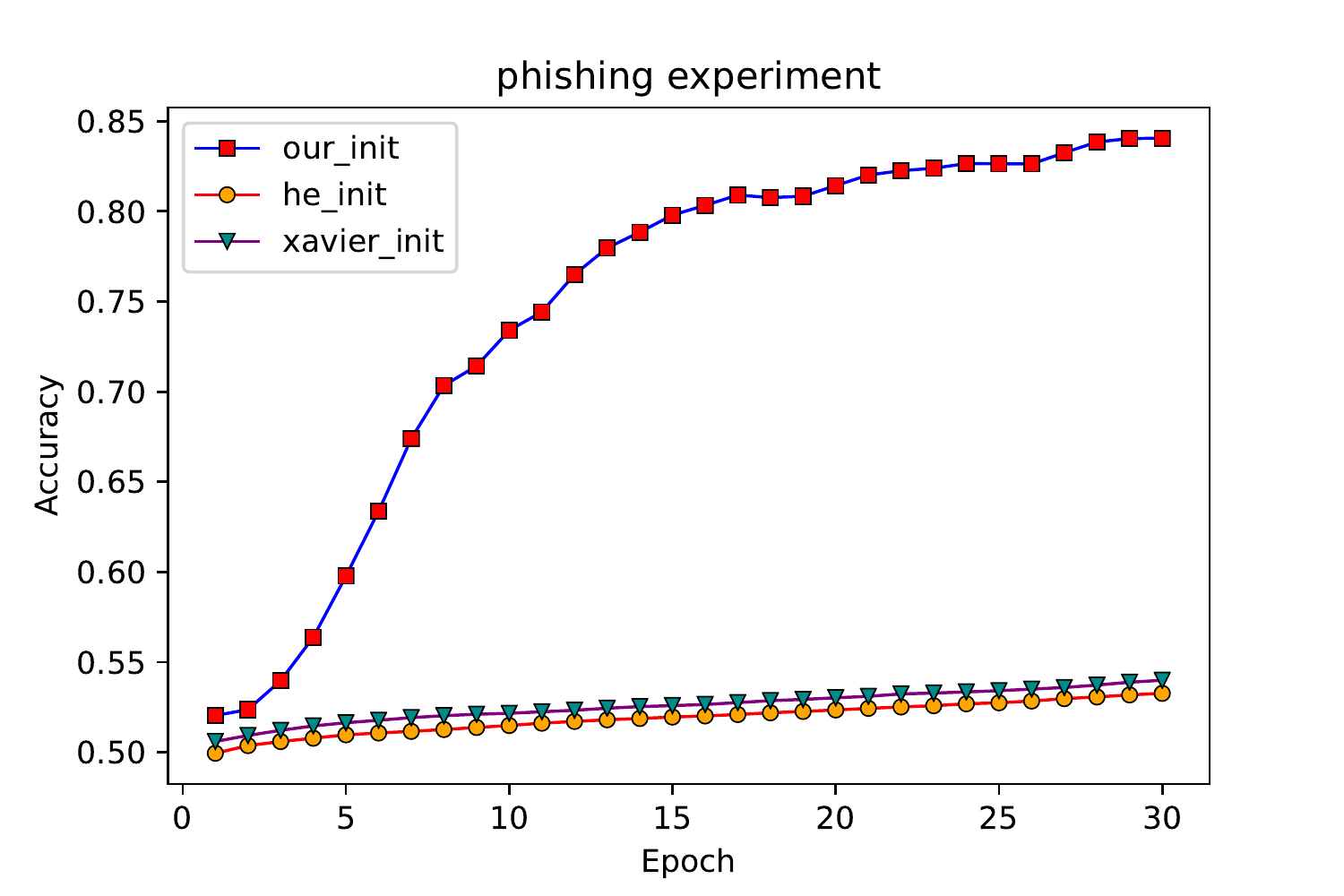}
  \caption{Initialization method comparing on Phishing Data Set}
  \label{fig: phishing_init_comparing}
\end{subfigure}
\begin{subfigure}{.49\textwidth}
  \centering
  \includegraphics[width=.9\linewidth]{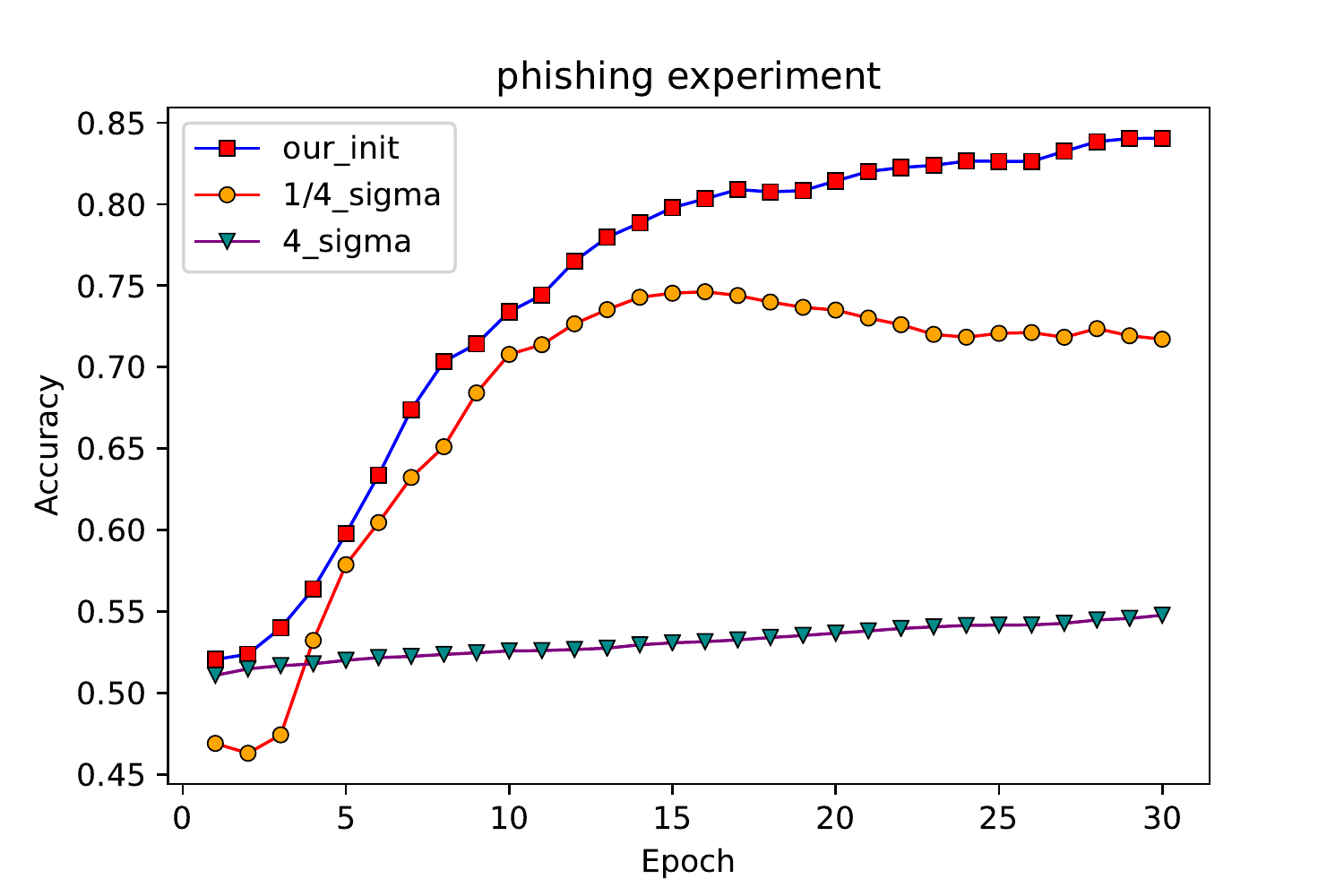}
  \caption{Initialization with $\sigma$, $\frac{1}{4}$ or $4\sigma$}
  \label{fig: phishing_init_std_comparing}
\end{subfigure}\\
\begin{subfigure}{.49\textwidth}
  \centering
  \includegraphics[width=.9\linewidth]{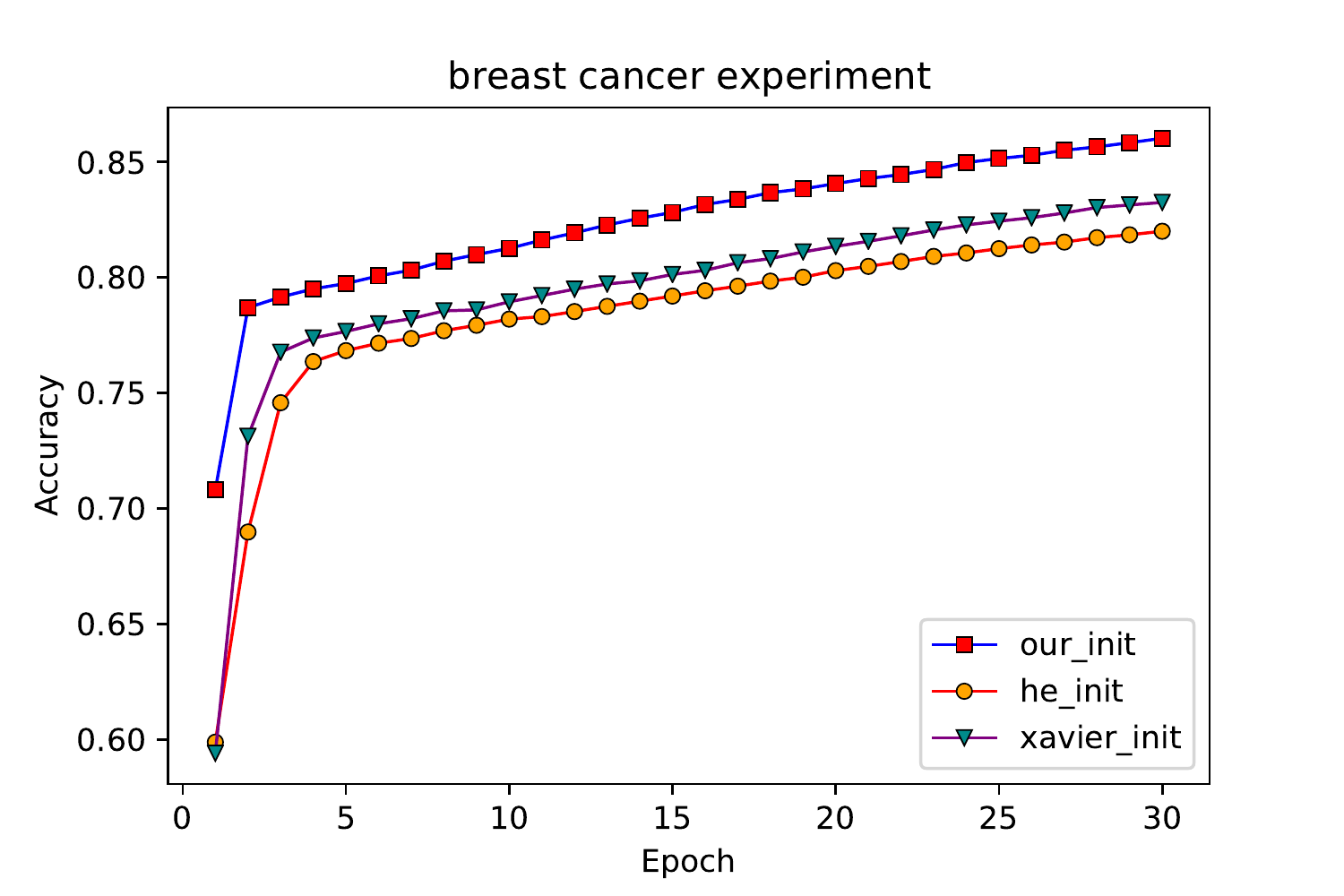}
  \caption{Initialization method comparing on Breast Cancer Data Set}
  \label{fig: breast_cancer_init_comparing}
\end{subfigure}
\begin{subfigure}{.49\textwidth}
  \centering
  \includegraphics[width=.9\linewidth]{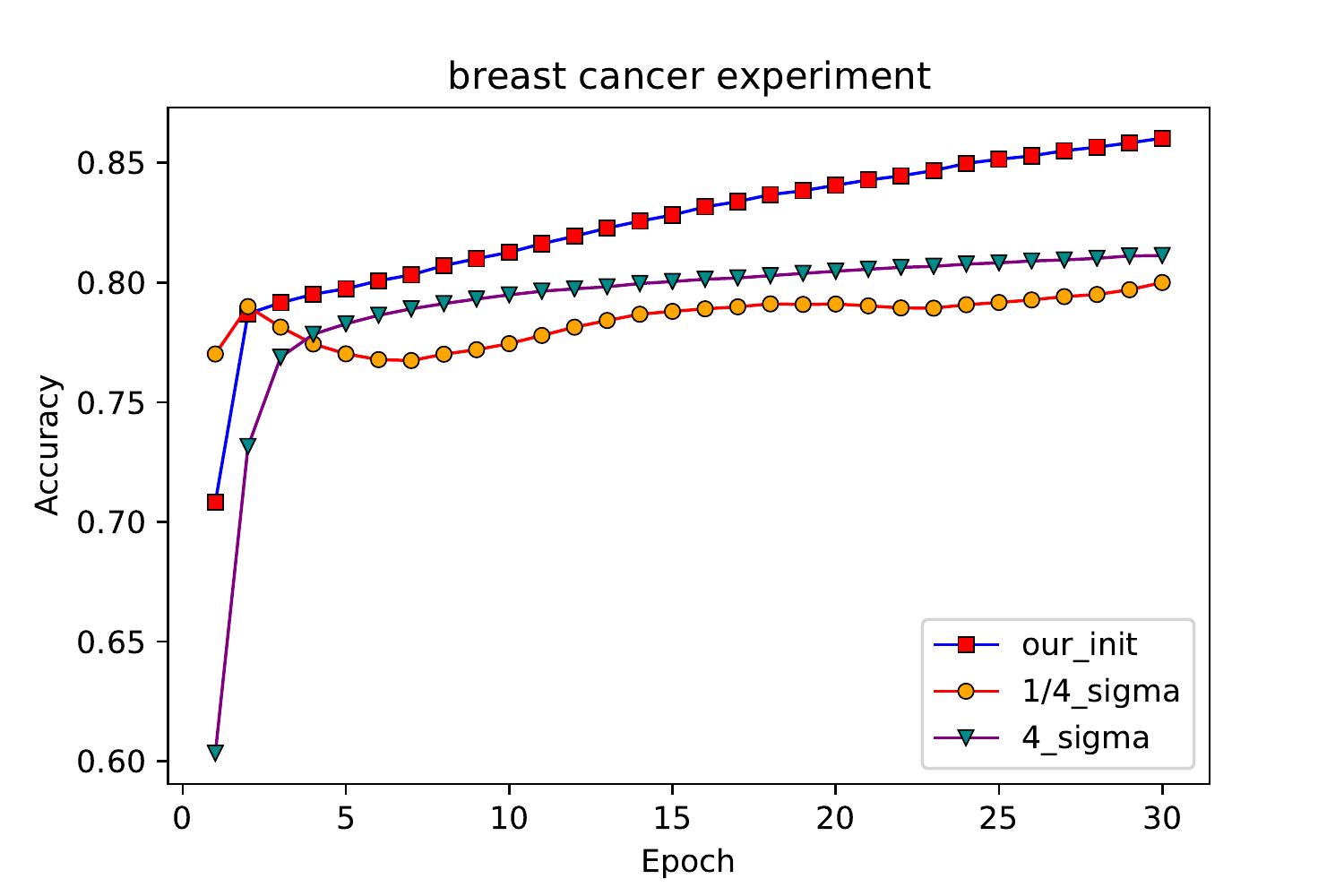}
  \caption{Initialization with $\sigma$, $\frac{1}{4}\sigma$ or $4\sigma$}
  \label{fig: breast_cancer_std_comparing}
\end{subfigure}
\label{fig: breast_cancer_experiments}
\caption{In (a), (c) and (e), we compare the accuracy curve with different initialization method on MNIST, Phishing and Breast Cancer Data set. In (b), (d) and (f), we test the convergence of the Bayesian Tensor Network model by perturbing the standard deviation  given by our formula.}
\label{fig: mps_init_experiments}
\end{figure}

In Fig. \ref{fig: mps_init_experiments}, we compare the accuracy on the MNIST Data set in the first $30$ epochs by our initialization method, Xavier initialization method and He initialization method. Our initialization method converges much more quickly than the other two initialization method on the MNIST data set. 

In Fig. \ref{fig: mnist_init_comparing}, Fig. \ref{fig: phishing_init_comparing} and Fig. \ref{fig: breast_cancer_init_comparing}, we compare the convergence of the Bayesian Tensor Network with our initialization method, Xavier initialization and He initialization method on the MNIST data set, Phishing data set and Breast Cancer data set. In these three data set, our initialization method works much better than the other two methods. Since the number of features in the Breast Cancer data set is only $10$, so the training of the Bayesian Tensor Network is not as the same heavily sensitive to the initialization method as the Phishing data set or the MNIST data set.

In Fig. \ref{fig: mnist_init_std_comparing}, Fig. \ref{fig: phishing_init_std_comparing}, and Fig. \ref{fig: breast_cancer_std_comparing}, we show the accuracy curves with the standard deviation $\sigma$ obtained by our formula, and slightly scaled std deviation. In the data set whose data has more number of features, the training process is more sensitive to the initialization and small deviation from the std given by our formula will lead to the bad convergence.

\section{Bayesian Framework For Tensor Networks}



%
%
%

\subsection{The General Framework}
We write down our Bayesian Tensor Networks model as follows, 
\begin{equation}
\begin{cases}
(\mathbf{A}|\alpha, \mathcal{B}) \sim \mathcal{N}(\mathbf{A}|\mathbf{0}, \alpha^{-1}\mathbf{I}),\\
(\mathbf{t}^{(i)}|\mathbf{x}^{(i)}, \mathbf{A})\sim 
p(\mathbf{t}^{(i)}|\mathbf{x}^{(i)}, \mathbf{A}).
\end{cases}
\end{equation}

In the inference step, we write down the posterior distribution of the parameters $\mathbf{A}$, namely $p(\mathbf{A}|\mathcal{D}, \mathcal{B})$ as 
\begin{align}
    p(\mathbf{A}|\mathcal{D}, \alpha, \mathcal{B}) \propto p(\mathcal{D}|\mathbf{A})p(\mathbf{A}|\alpha, \mathcal{B}). 
\end{align}

If we use the Maximum A Posterior Estimator (M.A.P.) to estimate the parameters in the optimal mode of the model which is the same as the normal Tensor Network model, 
\begin{align}
   \mathbf{A}_{\textit{MAP}} = \arg\max_{\mathbf{A}} \log{p(\mathbf{A}|\mathcal{D}, \alpha, \mathcal{B})}.
\end{align}

In our work, we focus on the Full Bayesian Analysis. We do prediction and make decision based on the predictive posterior marginal distribution. 
\begin{align}
    p(\hat{\mathbf{t}}|\hat{\mathbf{x}},\mathcal{D}, \alpha, \mathcal{B}) = \int{p(\hat{\mathbf{t}}|\mathbf{\hat{x}}, \mathbf{A})p(\mathbf{A}|\mathcal{D}, \alpha, \mathcal{B})d\mathbf{A}}
\end{align}

However, since the analytical intractability of the predictive posterior marginal distribution, we approximate the marginal predictive distribution around the M.A.P. mode of the posterior distribution till second order. We note that if we approximate the posterior distribution roughly by the Dirac distribution at the M.A.P. mode, 
\begin{align*}
    p(\mathbf{A}|\mathcal{D}, \alpha, \mathcal{B}) \simeq \delta(\mathbf{A}-\mathbf{A}_{\textit{MAP}}),
\end{align*}
we get
\begin{align}
    p(\hat{t}|\hat{x}, \mathcal{D}, \alpha, \mathcal{B}) \simeq p(\hat{t}|\hat{x}, \mathbf{A}_{\textit{MAP}}, \alpha, \mathcal{B}).
\end{align}
Obviously the information contained in the data set $\mathcal{D}$ is lost in the point approximation.

According to the predictive posterior marginal distribution $p(\mathbf{\hat{t}}|\mathbf{\hat{x}}, \mathcal{D}, \alpha, \mathcal{B})$ obtained in the prediction step, we can assign every new observation $\hat{\mathbf{x}}$ in the test set with one label $\mathbf{\hat{t}}$ which is a decision problem. For the goal of minimizing the misclassification rate in the classification problem, the decision boundary is determined as 
\begin{align}
    \mathcal{C} = \arg\max_{\mathcal{\hat{\mathbf{t}}}}p(\hat{\mathbf{t}}|\mathcal{D}, \alpha, \mathcal{B})
\end{align}

For some practical case where utility matrix (negative loss matrix) is specially designed, we can also maximize the expected utility function to determine the action. For the regression problem, we focus on the mean square loss and by minimize the expected loss, it is proved that the decision boundary is the conditional expectation of the label given the features, namely $\mathrm{E}_{\mathbf{t}}(\mathbf{t}|\mathbf{x})$, which is just the value of the regression function $y(\mathbf{A}, \mathbf{x})$.

\subsection{Classification}
In our convention, the data set is notated as 
\begin{align}
\mathcal{D} = \{(\mathbf{x}^{(i)}, \mathbf{t}^{(i)})|i\in(1,\cdots, m), \mathbf{x}^{(i)}\in\mathbb{R}^{(n)}, \mathbf{t}^{(i)}\in\{1,2, \cdots, k\}\}, 
\end{align}
in the classification problem. In our set up, $\mathbf{t}^{(i)}$ is encoded as the one-hot vector which represents which category the data $\mathbf{x}^{(i)}$ belongs to.

For the binary classification problem, the response of the Bayesian Tensor Network model is the logits, namely 
\begin{align}
    \log{\frac{p(\mathbf{t}^{(i)}|\mathbf{x}^{(i)}, \mathbf{A})}{1-p(\mathbf{t}^{(i)}|\mathbf{x}^{(i)}, \mathbf{A})}} =  \psi(\mathbf{x}^{(i)}, \mathbf{A}).
\end{align}
If we treat every component of the encoded vector $\mathbf{t}^{(i)}$ independently and model each component with above logit formula, then it can be easily extended to multi-classification case. 

In the multi-classification case, we use the Softmax activation function and we get 
\begin{align}
    p(\mathbf{t}^{(i)}_{j}|\mathbf{x}^{(i)}, \mathbf{A}) = \textit{softmax}(\psi_{j}(\mathbf{x}^{(i)}, \mathbf{A})).
\end{align}
For the binary case, we can write down the cost function as $-\log{p(\mathbf{A}|\mathcal{D}, \alpha, \mathcal{B})}$
\begin{align}
    \mathcal{L}(\mathbf{A}) \sim -\sum_{i}^{m} (t^{(i)}\log{\sigma(\psi(\mathbf{A}, \mathbf{x}^{(i)}))} + (1 - t^{(i)})\log{(1-\sigma(\psi(\mathbf{A}, \mathbf{x}^{(i)}))})) + \frac{1}{2\alpha}|\mathbf{A}|^{2}.
\end{align}
For the multi-classification case, we have 
\begin{align}
   \mathcal{L}(\mathbf{A}) \sim -\sum_{ij}(\psi^{(i)}_{j} + \log{(\sum_{j}\psi_{j}^{(i)}})) + \frac{1}{2\alpha}|\mathbf{A}|^{2}.
\end{align}

In the Bayesian Tensor Network, we do not need to do inference (training) if we can solve the intractable posterior marginal distribution as long as the the prior distribution is wisely introduced according to the background knowledge. However, we need to find the M.A.P mode of the posterior distribution to expand the posterior distribution around the M.A.P mode. We write down the posterior distribution and use the stochastic gradient optimization method to get to the M.A.P. mode. In practice, our objective function is the negative log posterior distribution. Around the optimal mode, we get the normal distribution as  
\begin{align}
    p(\mathbf{A}|\mathcal{D}, \alpha, \mathcal{B}) \simeq q(\mathbf{A}|\mathcal{D}, \alpha, \mathcal{B}), 
\end{align}
where 
\begin{align*}
    q(\mathbf{A}|\mathcal{D}, \alpha, \mathcal{B}) = \mathcal{N}(\mathbf{A}|\mathbf{A}_{\textit{MAP}}, \mathbf{M}^{-1}).
\end{align*}

The Hessian matrix contains the geometric information (curvature) of the posterior distribution, so more information is extracted by the Hessian matrix which is a better approximation than the $\delta(\mathbf{A}-\mathbf{A}_{\textit{MAP}})$ distribution.

The co-variance matrix of the Normal distribution $q(\mathbf{A}|\mathcal{D}, \alpha, \mathcal{B})$ is 
\begin{align}
\mathbf{M} &= \frac{\partial^{2}}{\partial{\mathbf{A}^{2}}}\log{p(\mathbf{A}|\mathcal{D}, \alpha, \mathcal{B})} = H + \frac{1}{\alpha}\mathrm{I}, 
\end{align}
where $H$ is the second derivative matrix  of the log likelihood function $\log{p(\mathcal{D}|\mathbf{A}, \alpha, \mathcal{B})}$. The time complexity of computing the inverse of the Hessian matrix $M$ is $O(n^{3})$ which is time consuming, so we use the Out-Product approximation to decrease the time complexity to $O(n)$.

Since the prior distribution $p(\mathbf{A}|\alpha)$ and the approximated posterior distribution $q(\mathbf{A}|\mathcal{D}, \mathcal{B})$ are all Normal distribution, then we can get the predictive marginal posterior distribution analytically as 
\begin{align}
    p(\mathbf{\hat{t}}|\mathbf{\hat{x}}, \mathcal{D}, \alpha) &=  \int{p(\mathbf{\hat{t}}|\mathbf{\hat{x}}, \mathbf{A})p(\mathbf{A}| \mathcal{D}, \mathcal{B})d\mathbf{A}}\\
    &\simeq \int{p(\mathbf{\hat{t}}|\mathbf{\hat{x}}, \mathbf{A})q(\mathbf{A}| \mathcal{D}, \mathcal{B})d\mathbf{A}}
\end{align}
By plugging in the approximated posterior distribution $q(\mathbf{A}|\mathcal{D}, \alpha, \mathcal{B})$, we get
\begin{align}
    p(\hat{t}_{j}|\mathbf{\hat{x}}, \mathcal{D}, \alpha) = \sigma(\kappa(\sigma^{2}_{j})\mu^{\prime}_{j}), 
\end{align}
where 
\begin{align*}
    &\sigma_{j}^{2} = \nabla{\psi^{j}_{\textit{MAP}}}\mathbf{M}^{-1}\nabla{\psi^{j}_{\textit{MAP}}},\\
    &\mu_{j}^{\prime} = \psi^{j}_{\textit{MAP}} - \log{\sum_{-j}\exp{(\psi^{j}_{\textit{MAP}})}}.
\end{align*}

\subsection{Hessian Matrix}
The approximation of the Hessian matrix has been widely studied. In the Out-Product approximation method, the idea is in the trained networks, the label $t^{(i)}_{j}$ and the output $\hat{y}^{(i)}_{j}$ are close to each other then the second derivative matrix term is very small which is ignored. We get the out-product approximation of the Hessian matrix of the Bayesian Tensor Network model as 
\begin{align}
   H = \sum_{i, j}t^{(i)}_{j}(\sum_{k, k^{\prime}}y^{(i)}_{k}y^{(i)}_{k^{\prime}}\frac{\partial{\psi^{(i)}_{k}}}{\partial{\mathbf{A}}}\frac{\partial{\psi^{(i)}_{k^{\prime}}}}{\partial{\mathbf{A}}}- \sum_{k}y_{k}^{(i)}(\frac{\partial{\psi^{(i)}_{k}}}{\partial{\mathbf{A}}})^{2})
\end{align}

Our result above just contains the first derivative which means the time complexity is almost $O(n)$. Here $\psi^{(i)}_{k}$ means the $k$ component of the output of the Bayesian Tensor Networks. Different from the Neural Networks, the first derivative of the logits can be calculated analytically and then we can get an analytical result of the Hessian matrix of the Bayesian Tensor Networks. 

%

\subsection{Numerical Results} 
We study the performance of the Bayesian Tensor Networks on several data set. 


From small data set to big data set, we used the following data sets

\begin{itemize}
\item Synthetic Data Set: Two dimensional Gaussian Blobs with two classes.

\item Breast Cancer Wisconsin Data Set: A toy binary classification data set. 

\item Phishing Website Data Set: A small binary classification data set.

\item MNIST Data Set: A standard multi-classification data set in computer vision community.
\end{itemize}

To study the Bayesian effects, we visualize the parameters in the Bayesian Tensor Network and the decision boundary in two dimensional synthetic data set.

We study the performance of the Bayesian Tensor Network with different standard deviation and different bond dimension on the Breast Cancer Wisconsin, Phishing Website and MNIST data set.

\begin{figure}
\begin{subfigure}{.33\textwidth}
  \centering
  \includegraphics[width=.9\linewidth]{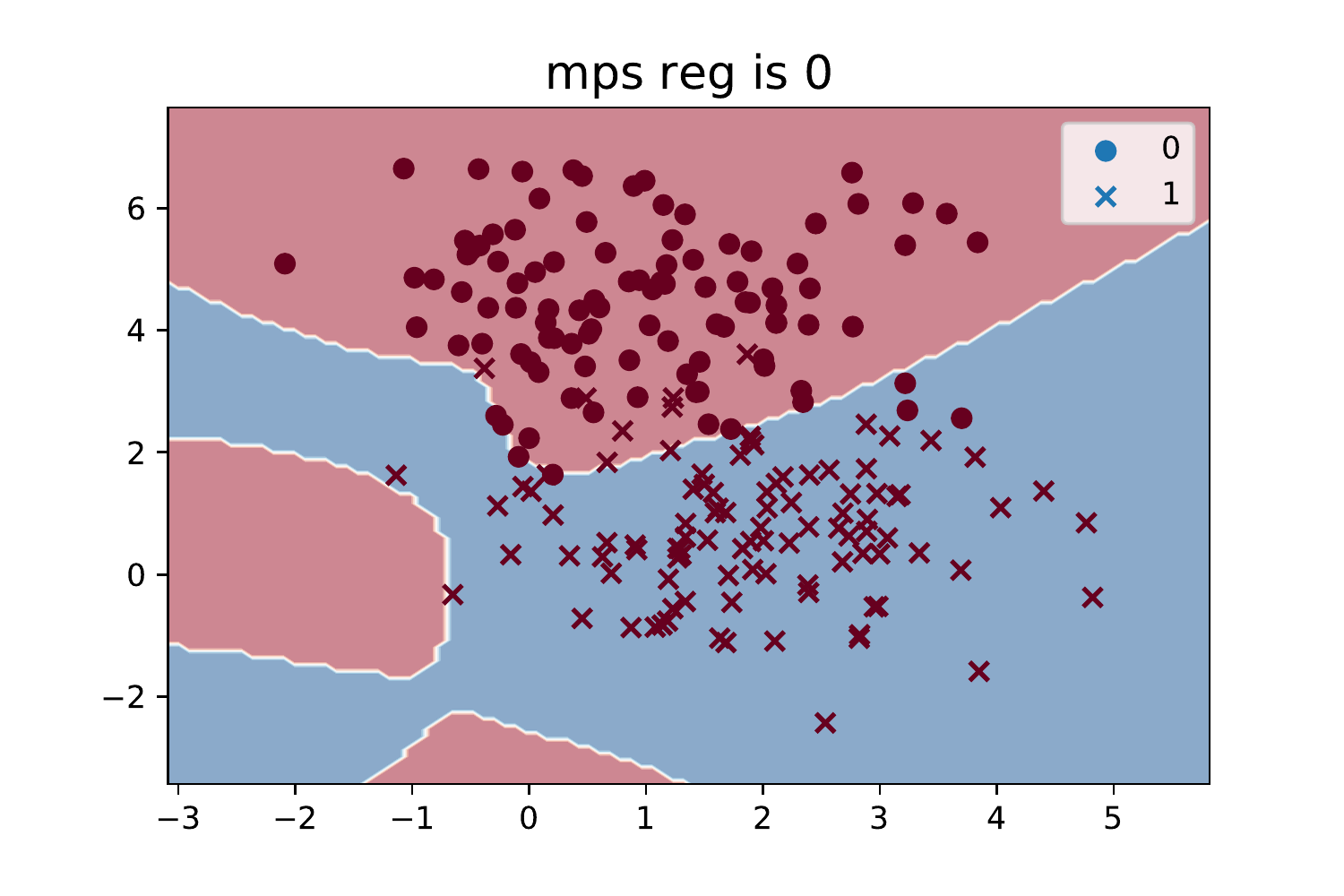}
  \caption{Normal MPS model}
  \label{fig: mps_reg_0_db_plot}
\end{subfigure}
\begin{subfigure}{.33\textwidth}
  \centering
  \includegraphics[width=.9\linewidth]{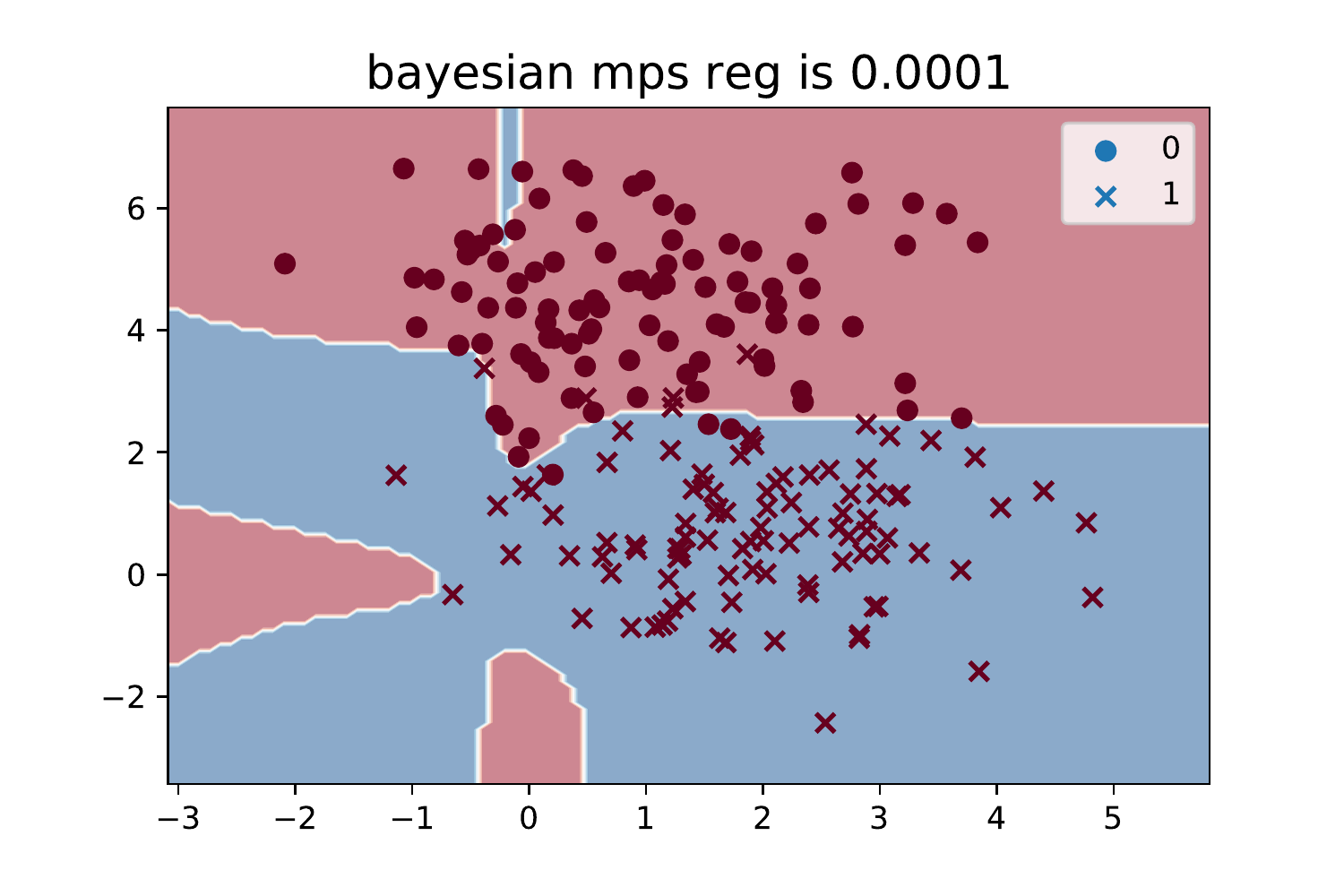}
  \caption{Bayesian Tensor Network with reg $0.0001$}
  \label{fig: mps_reg_0.0001_db_plot}
\end{subfigure}
\begin{subfigure}{.33\textwidth}
  \centering
  \includegraphics[width=.9\linewidth]{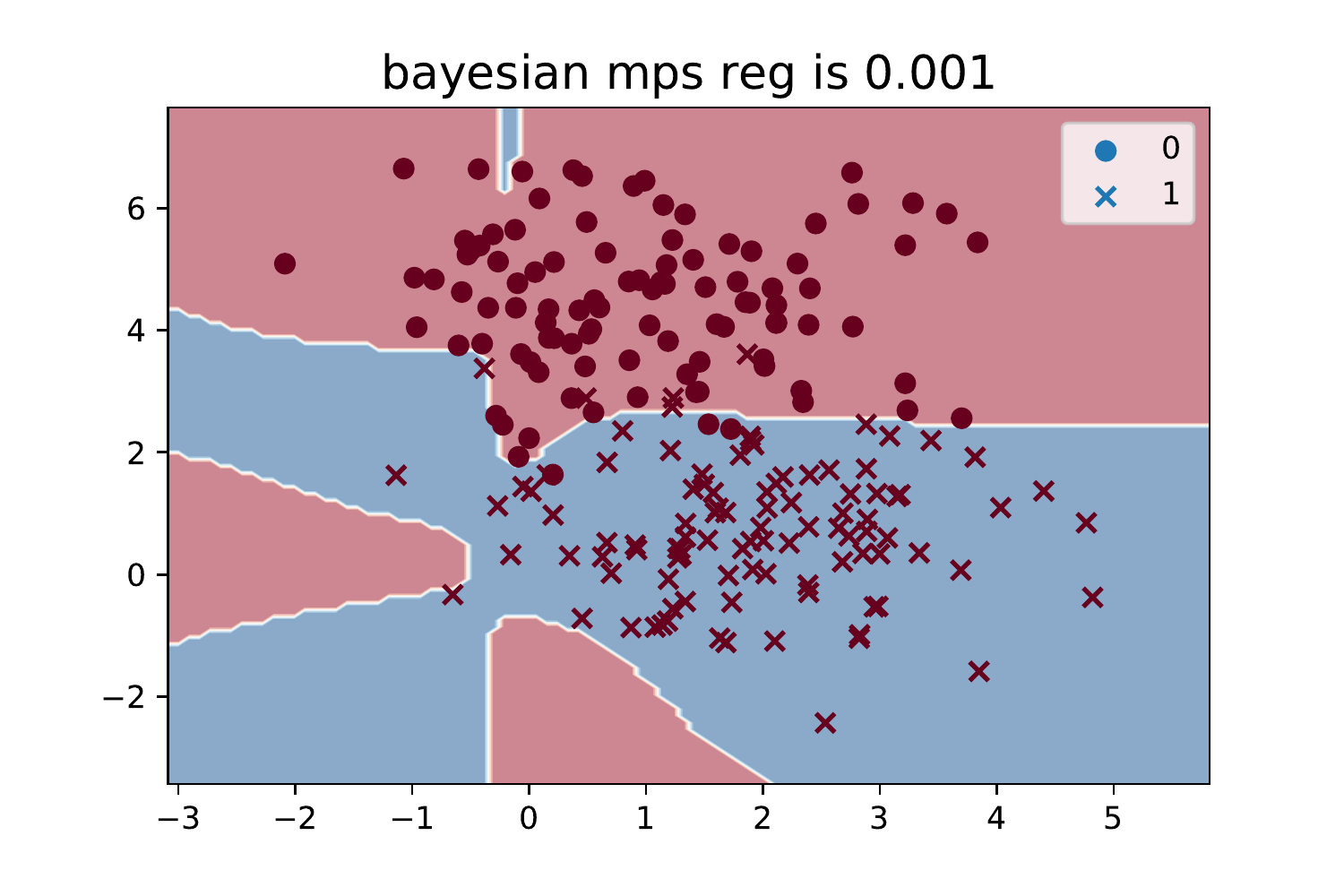}
  \caption{Bayesian Tensor Network with reg $0.001$}
  \label{fig: mps_reg_0.001_db_plot}
\end{subfigure}\\
\begin{subfigure}{.33\textwidth}
  \centering
  \includegraphics[width=.9\linewidth]{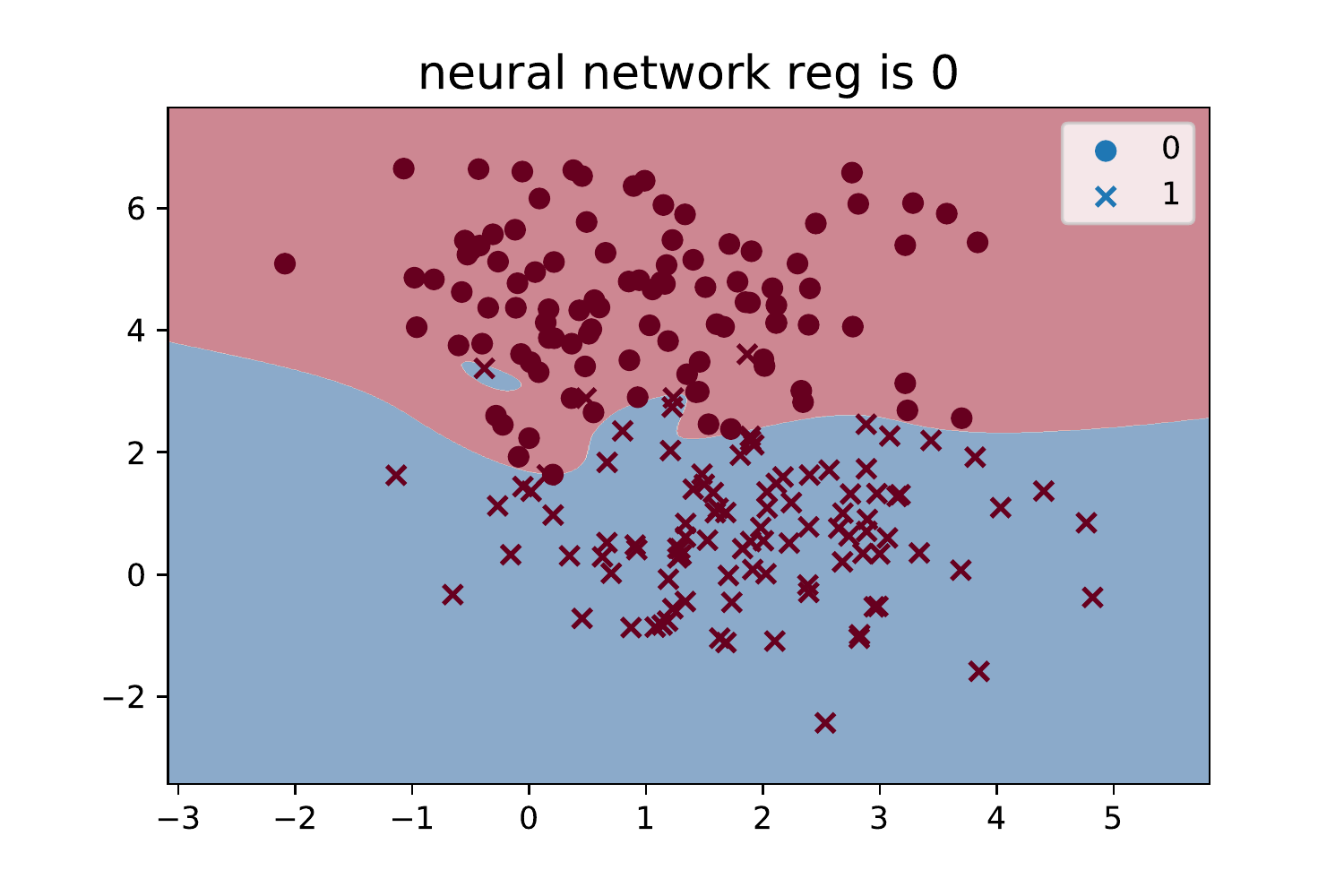}
  \caption{Normal Neural Network}
  \label{fig: nn_db_plot}
\end{subfigure} 
\begin{subfigure}{.33\textwidth}
  \centering
  \includegraphics[width=.9\linewidth]{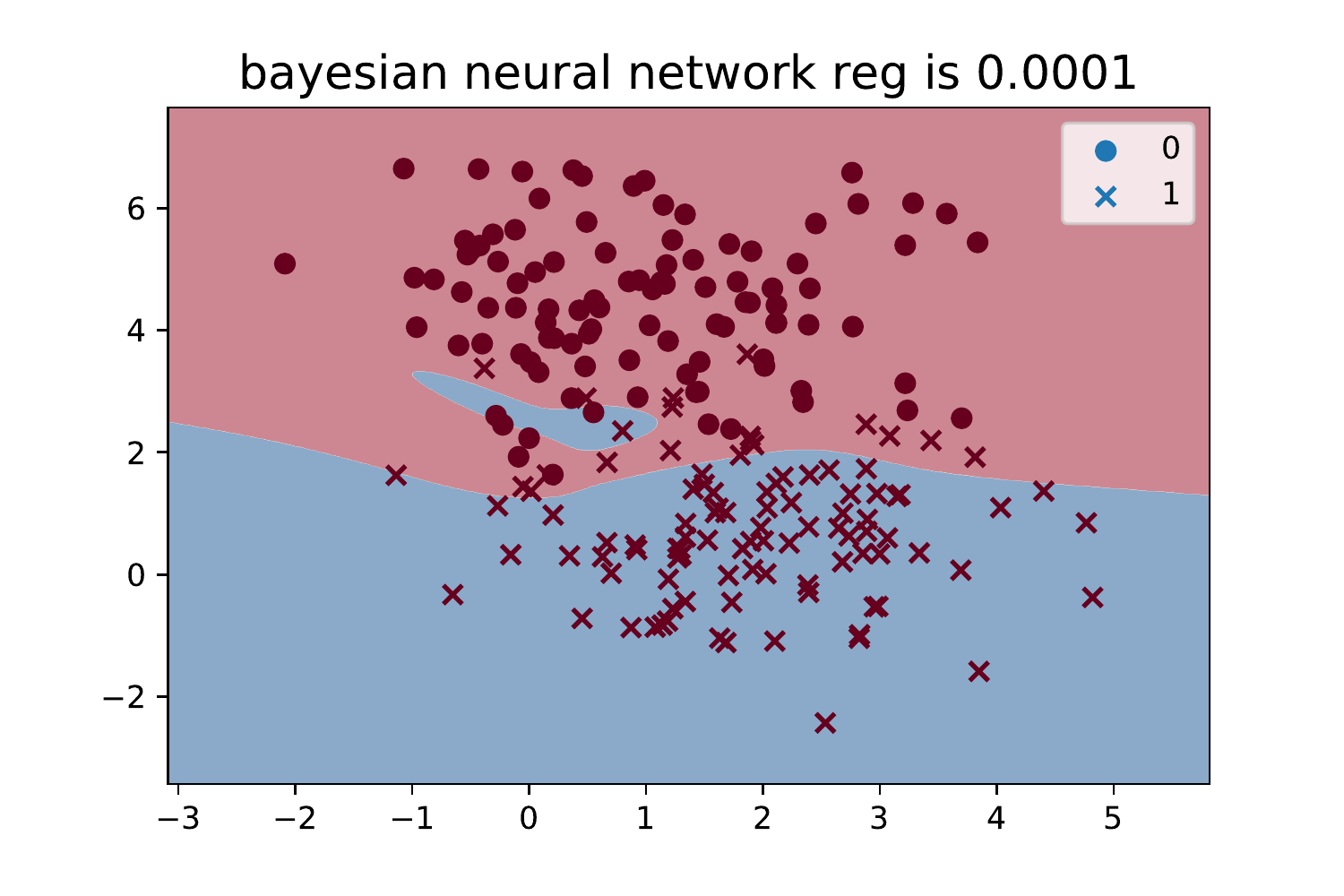}
  \caption{Bayesian Neural Network with reg $0.0001$}
  \label{fig: bnn_reg_0.0001_db_plot}
\end{subfigure} 
\begin{subfigure}{.33\textwidth}
  \centering
  \includegraphics[width=.9\linewidth]{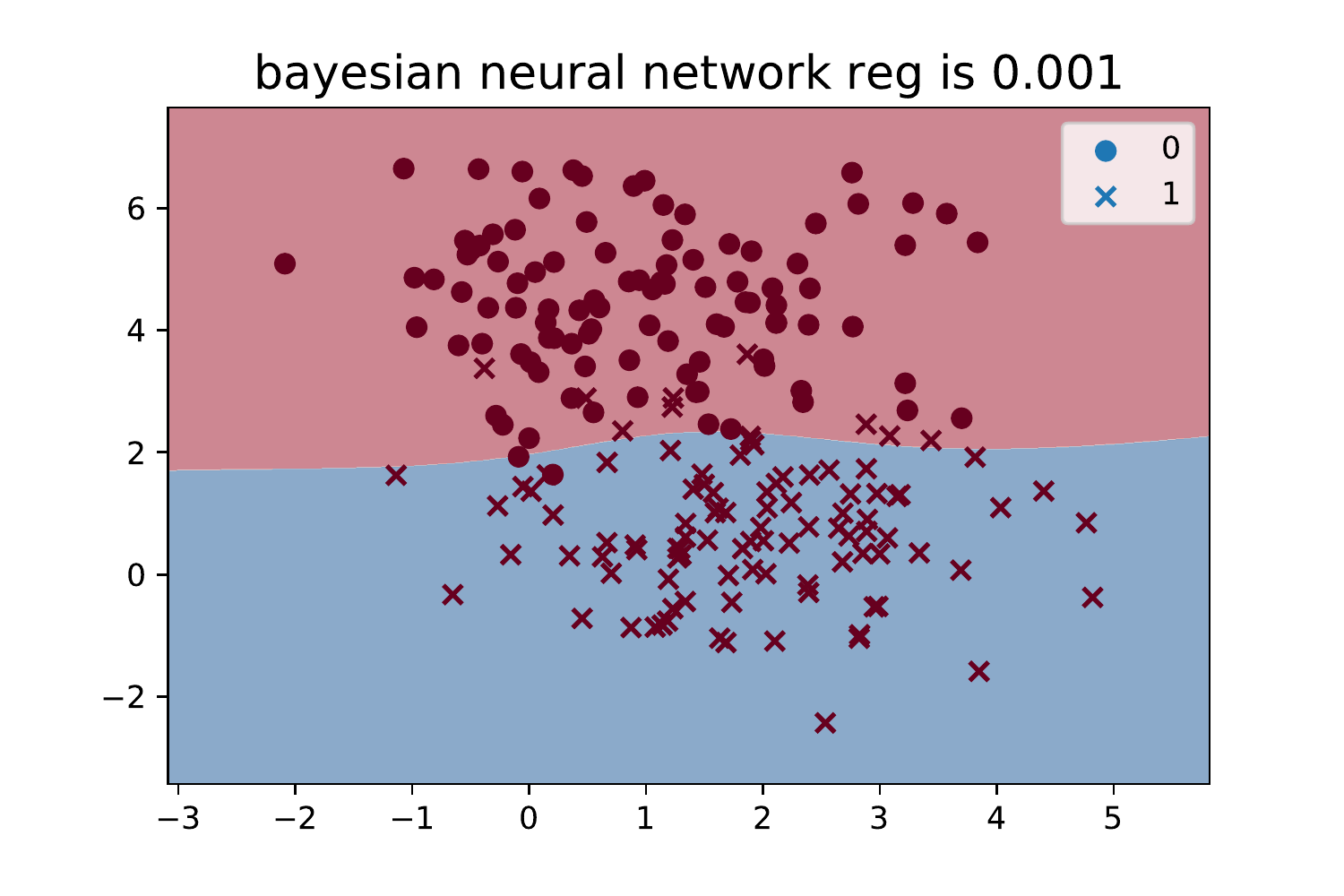}
  \caption{Bayesian Neural Network with reg $0.001$}
  \label{fig: bnn_reg_0.001_db_plot}
\end{subfigure}
\caption{In (a), (b), (c), we show the decision boundary of the Bayesian Tensor Network. In (d), (e), (f), we show the decision boundary of the Bayesian Neural Network.}
\label{fig: mps_bnn_db_plot}
\end{figure}

We train the Bayesian Tensor Networks and Bayesian Neural Networks on the blobs synthetic data set which contains $200$ samples in two classes. We used relatively bigger tensor nets and neural nets model to overfit the data set to study the Bayes shrinkage effect with different prior distribution in Bayesian Tensor Network and Bayesian Neural Network in Fig. \ref{fig: mps_bnn_db_plot}. 
As we use greater standard deviation in the prior Normal distribution, the decision boundary becomes smoother.
From our numerical experiments, we find that neural network is slightly more sensitive to the prior distribution than the tensor network. 

\begin{figure}
\begin{subfigure}{.33\textwidth}
  \centering
  \includegraphics[width=.9\linewidth]{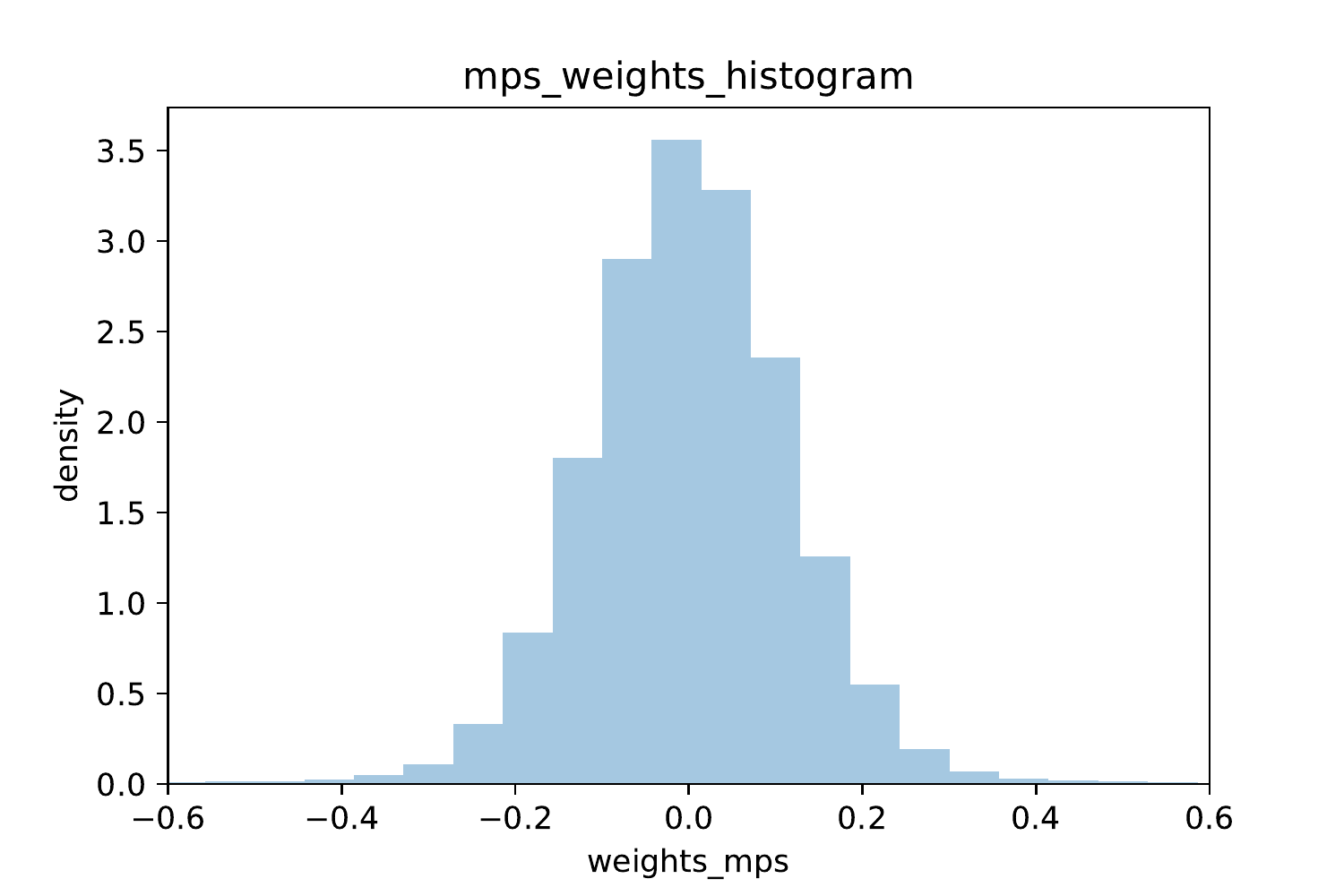}
  \caption{Normal Tensor Network}
  \label{fig: mps_reg_0_hist_plot}
\end{subfigure}
\begin{subfigure}{.33\textwidth}
  \centering
  \includegraphics[width=.9\linewidth]{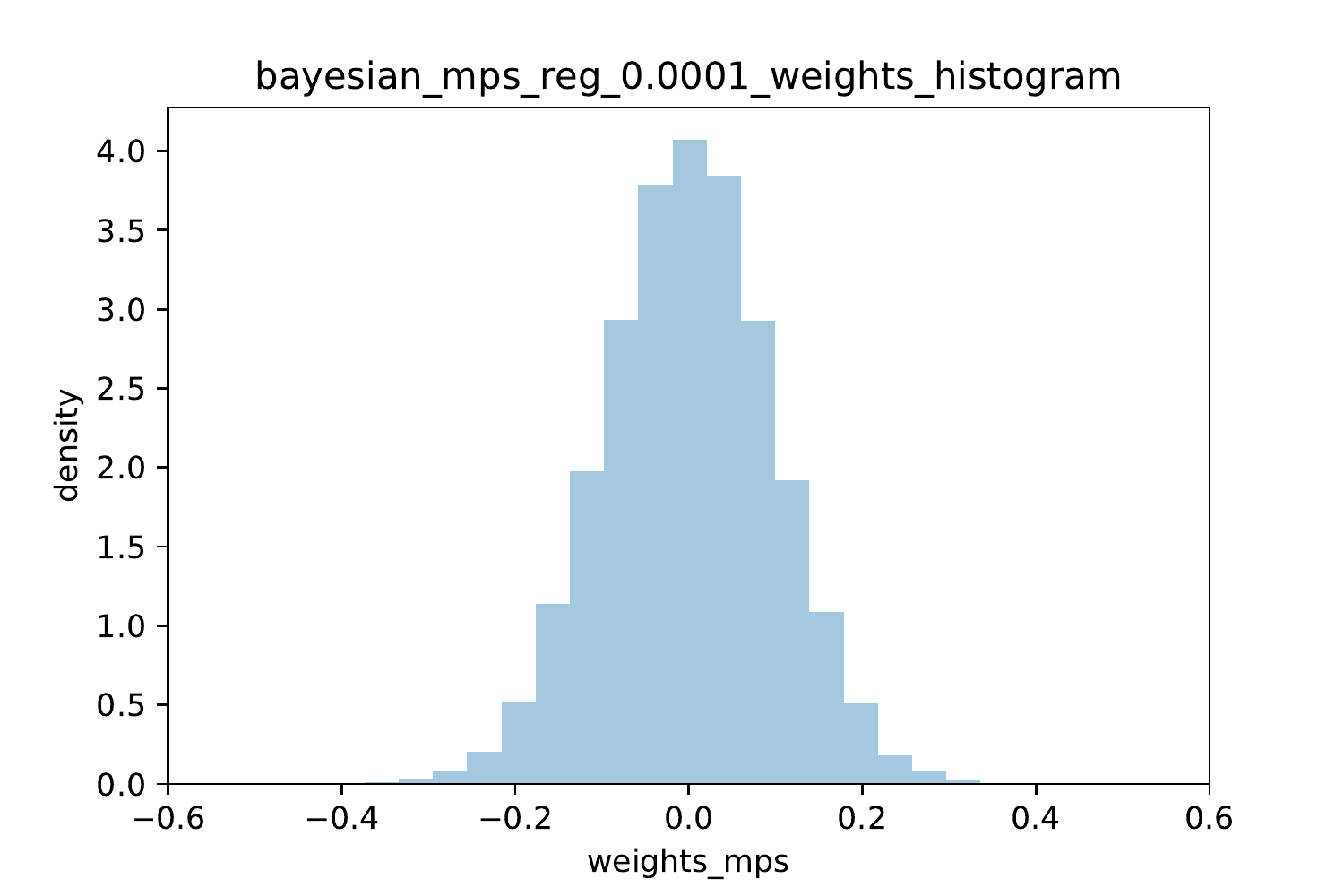}
  \caption{Bayesian Tensor Network with reg $0.0001$}
  \label{fig: mps_reg_0.0001_hist_plot}
\end{subfigure}
\begin{subfigure}{.33\textwidth}
  \centering
  \includegraphics[width=.9\linewidth]{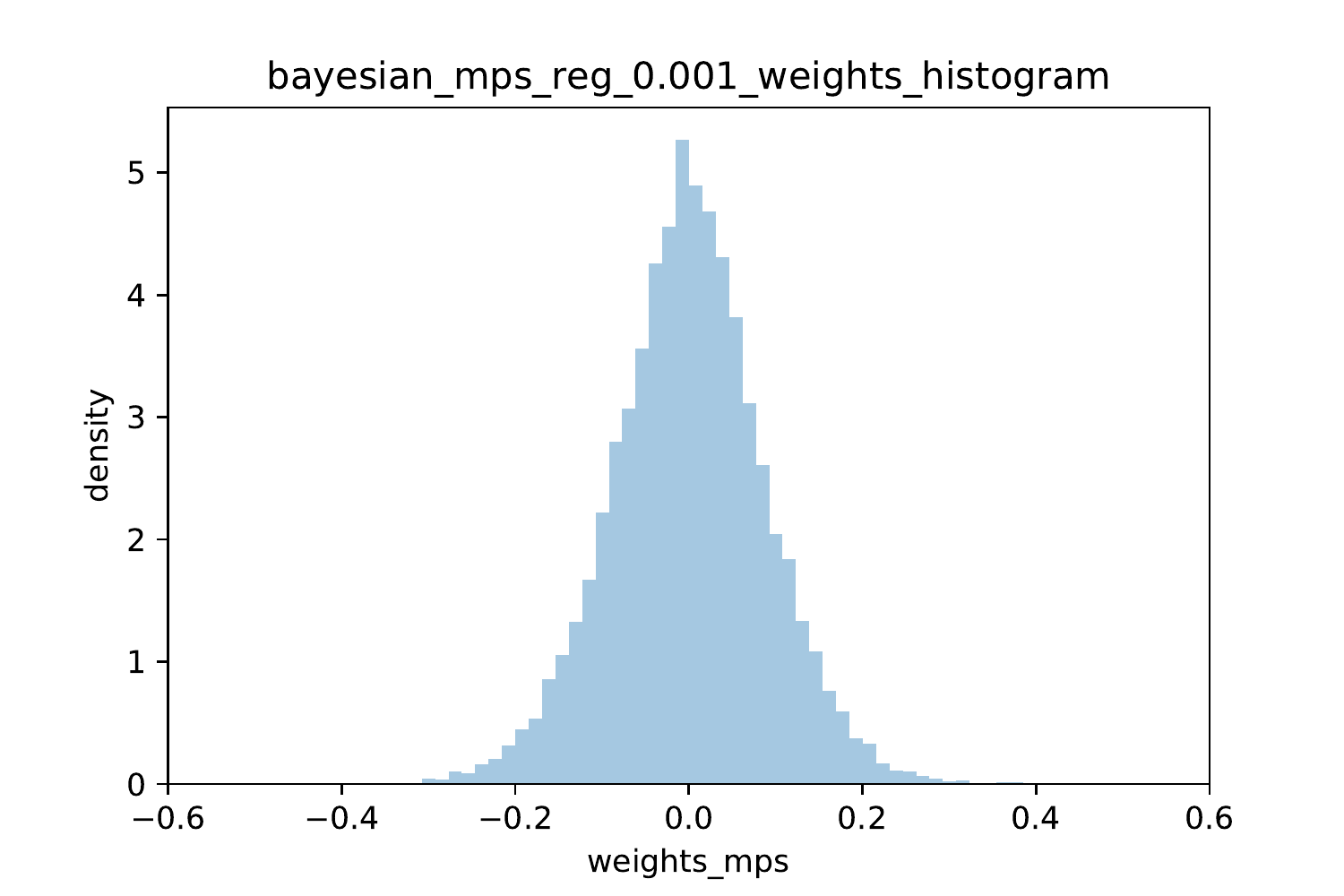}
  \caption{Bayesian Tensor Network with reg $0.001$}
  \label{fig: mps_reg_0.001_hist_plot}
\end{subfigure}\\
\begin{subfigure}{.33\textwidth}
  \centering
  \includegraphics[width=.9\linewidth]{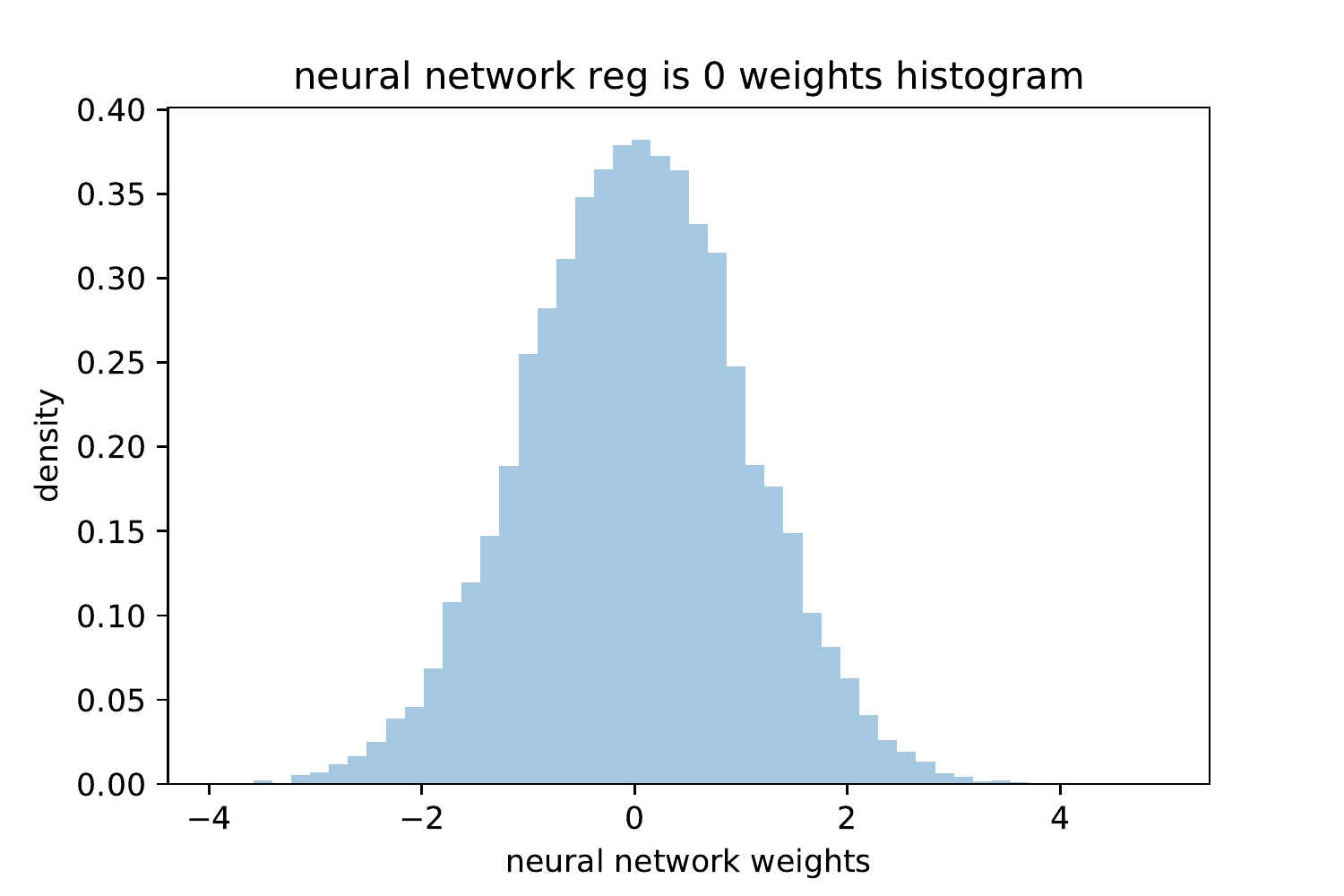}
  \caption{Normal Neural Network}
  \label{fig: nn_hist_plot}
\end{subfigure}
\begin{subfigure}{.33\textwidth}
  \centering
  \includegraphics[width=.9\linewidth]{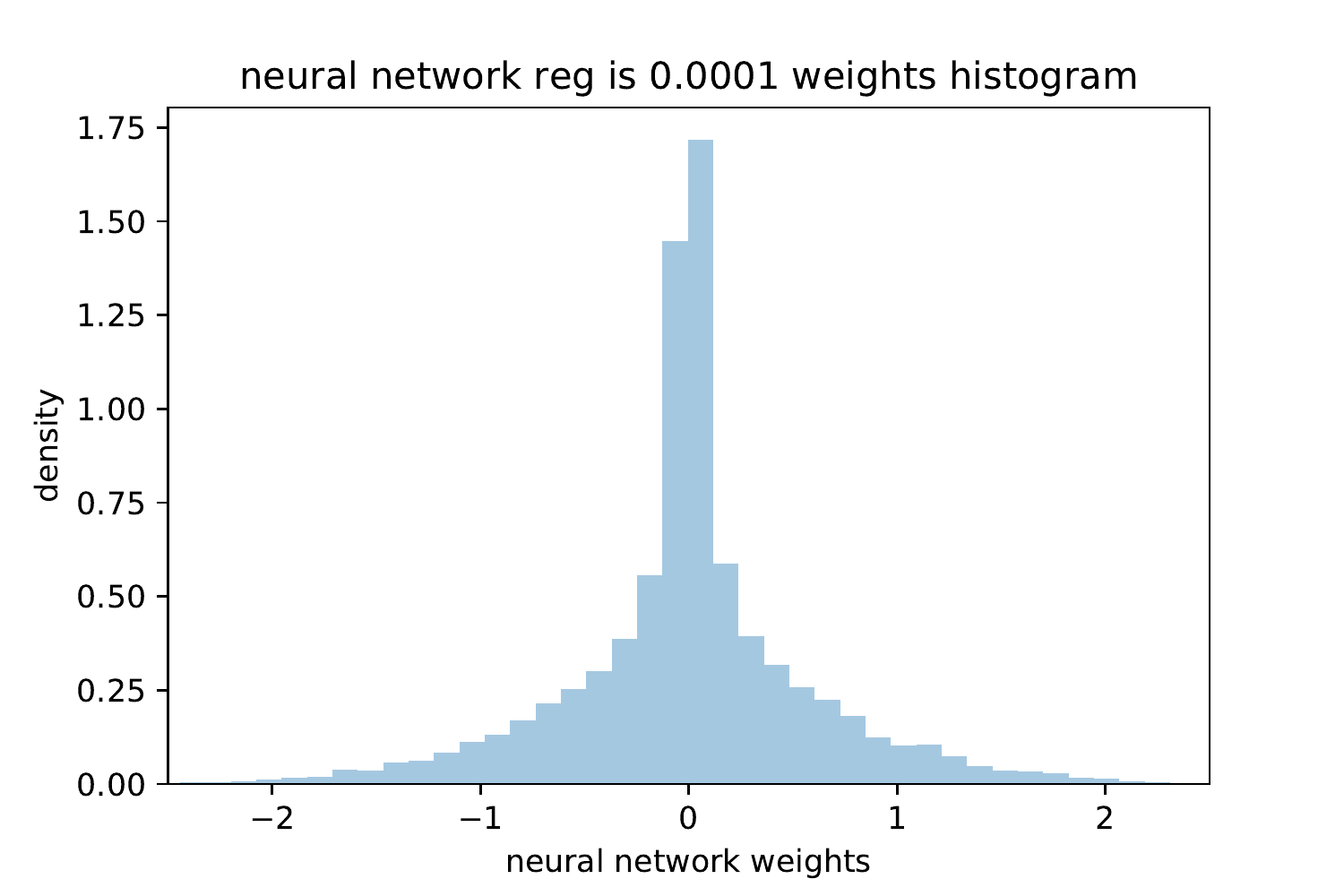}
  \caption{Bayesian Neural Network with reg $0.0001$}
  \label{fig: bnn_reg_0.0001_hist_plot}
\end{subfigure}
\begin{subfigure}{.33\textwidth}
  \centering
  \includegraphics[width=.9\linewidth]{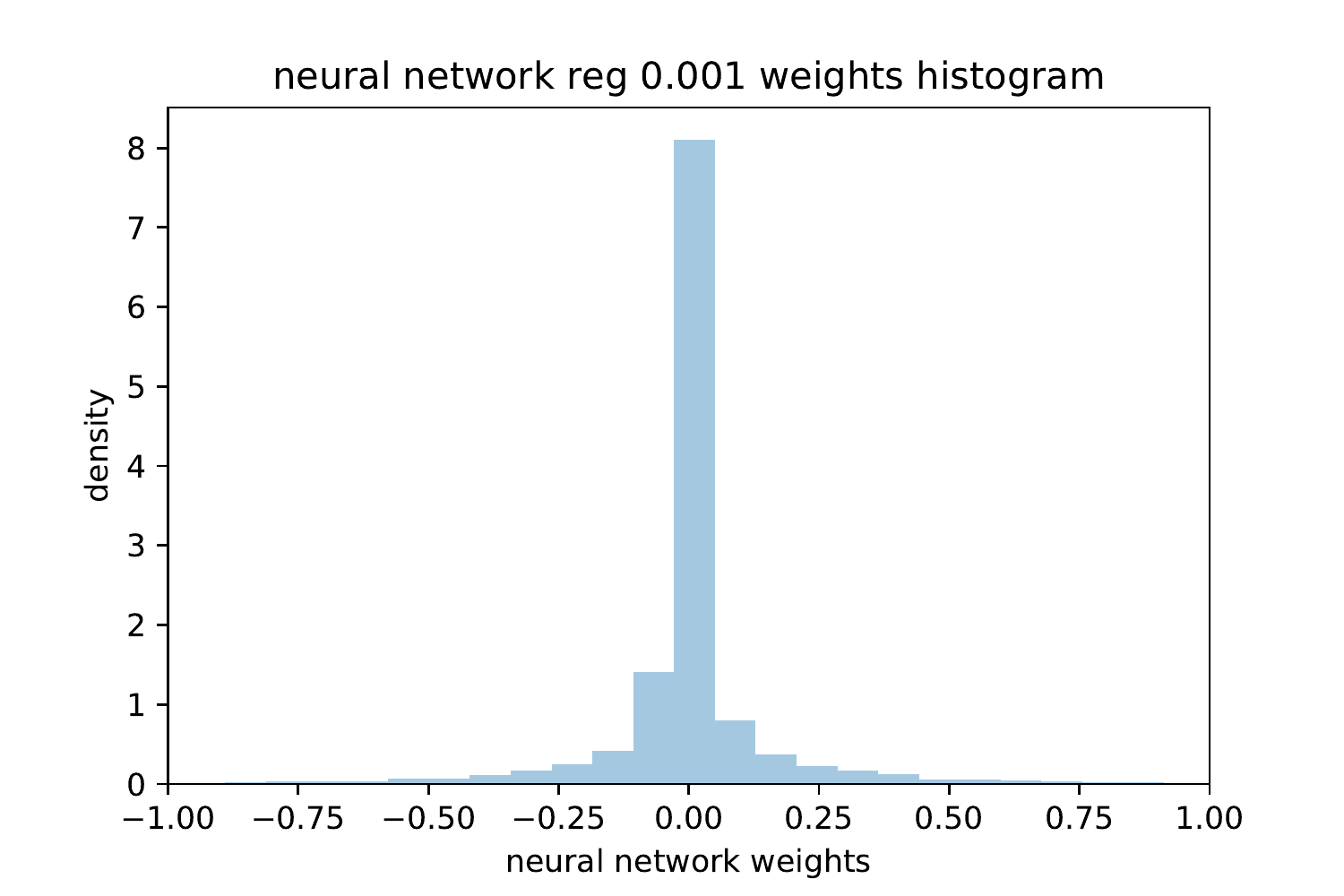}
  \caption{Bayesian Neural Network with reg $0.001$}
  \label{fig: bnn_reg_0.001_hist_plot}
\end{subfigure}
\caption{In (a), (b) and (c), we show the histogram of the parameters in Bayesian Tensor Network. In (c), (d), (e), we show the histogram of the parameters in Bayesian Neural Network.}
\label{fig: mps_bnn_hist_plot}
\end{figure}

In Fig. \ref{fig: mps_bnn_hist_plot}, we show the histogram of the parameters in the trained Bayesian Tensor Network and Bayesian Neural Network. We observe the Bayesian shrinkage in both the histograms of the parameters in the Bayesian Tensor Network and the Bayesian Neural Network. In the Bayesian Tensor Network, we find that the distribution of the parameters is not heavily affected instead the stand deviation is decreased as the prior std deviation is decreased. For the Bayesian Neural Network, we find that the Bayesian shrinkage effect is heavier and the parameters distribution gets to be heavy tail which means the parameters in the model become sparse.

\begin{figure}
\begin{subfigure}{.33\textwidth}
  \centering
  \includegraphics[width=.9\linewidth]{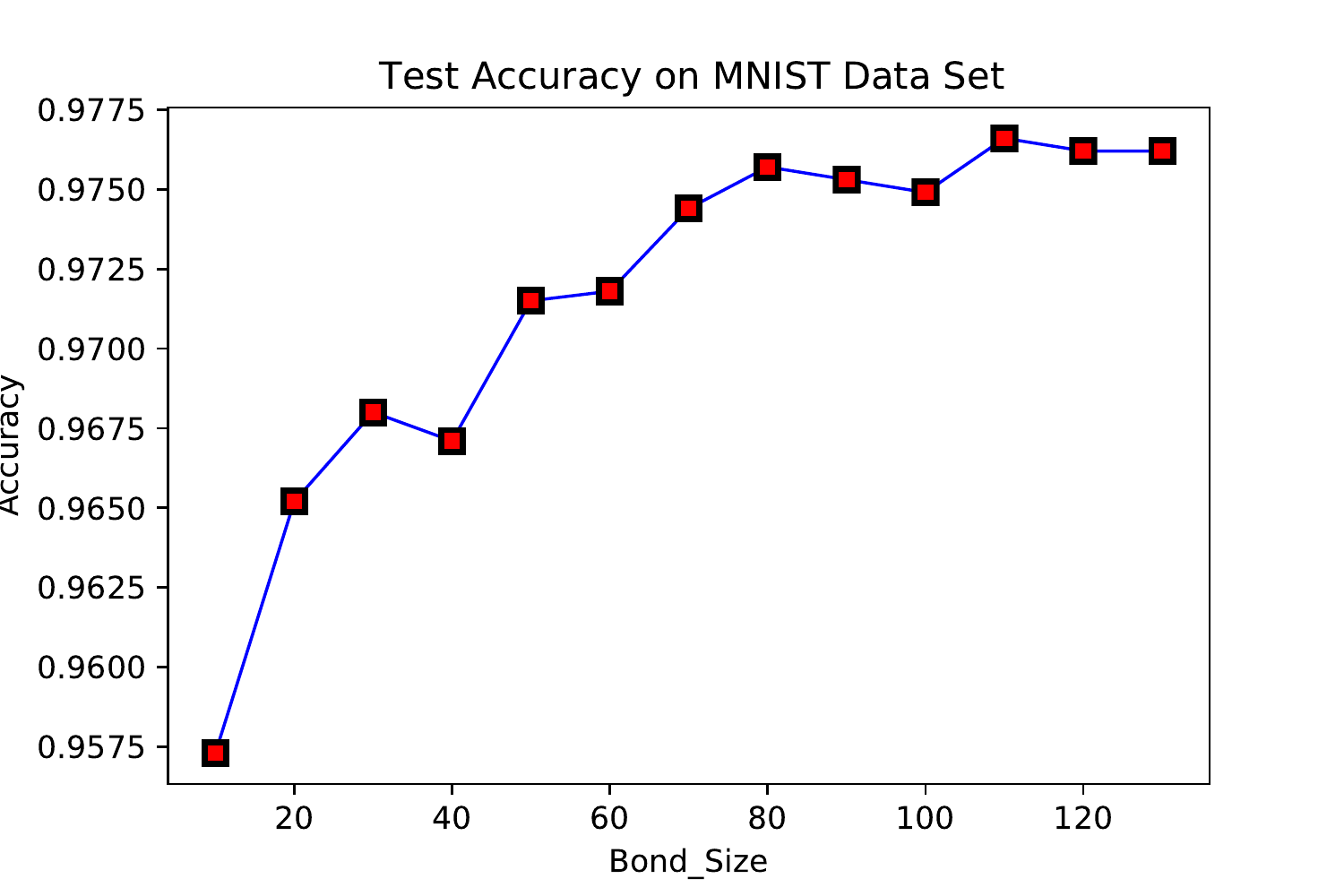}
  \caption{MNIST data set}
  \label{fig: acc_mnist_bond}
\end{subfigure} 
\begin{subfigure}{.33\textwidth}
  \centering
  \includegraphics[width=.9\linewidth]{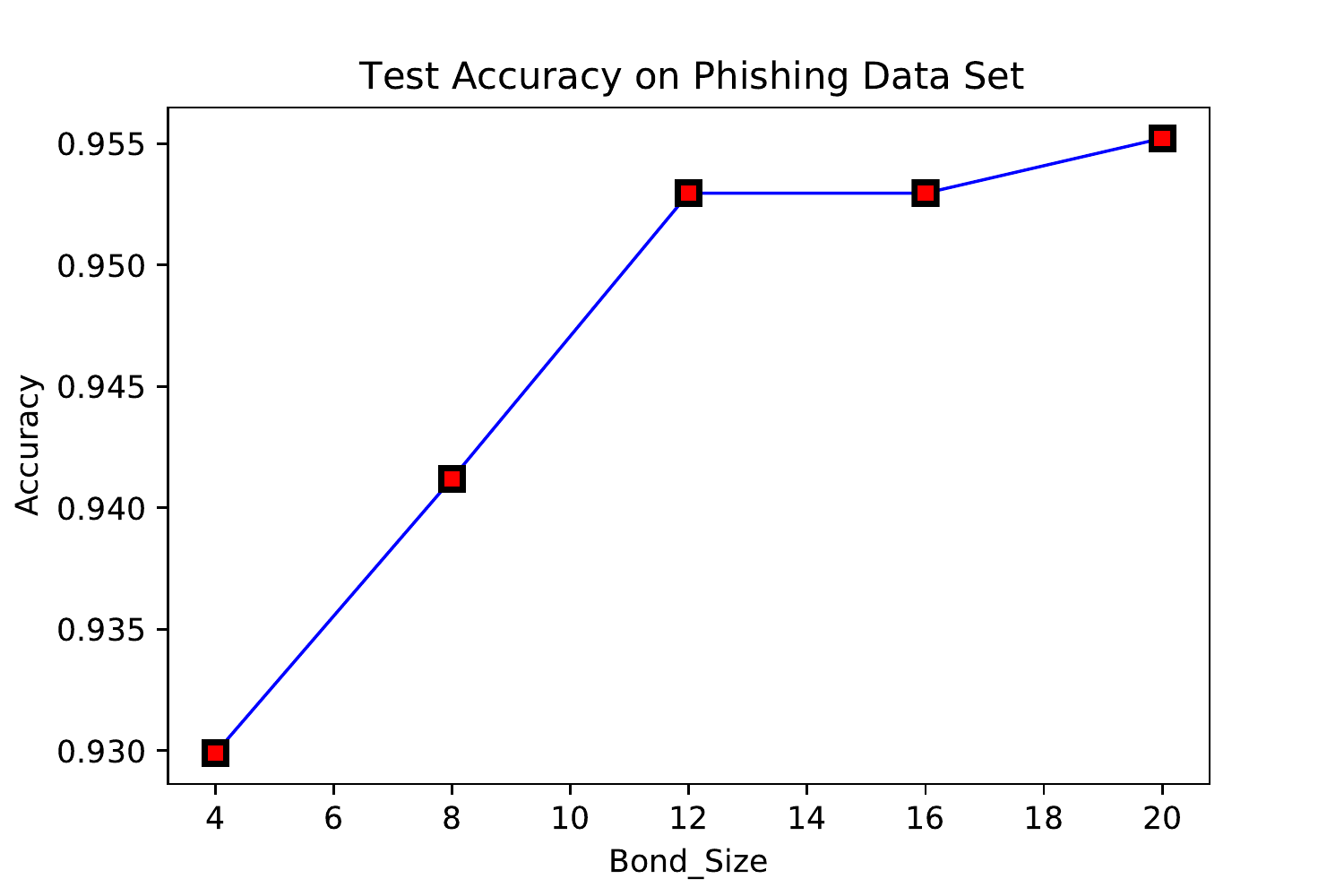}
  \caption{Phishing data set}
  \label{fig: acc_phishing_bond}
\end{subfigure} 
\begin{subfigure}{.33\textwidth}
  \centering
  \includegraphics[width=.9\linewidth]{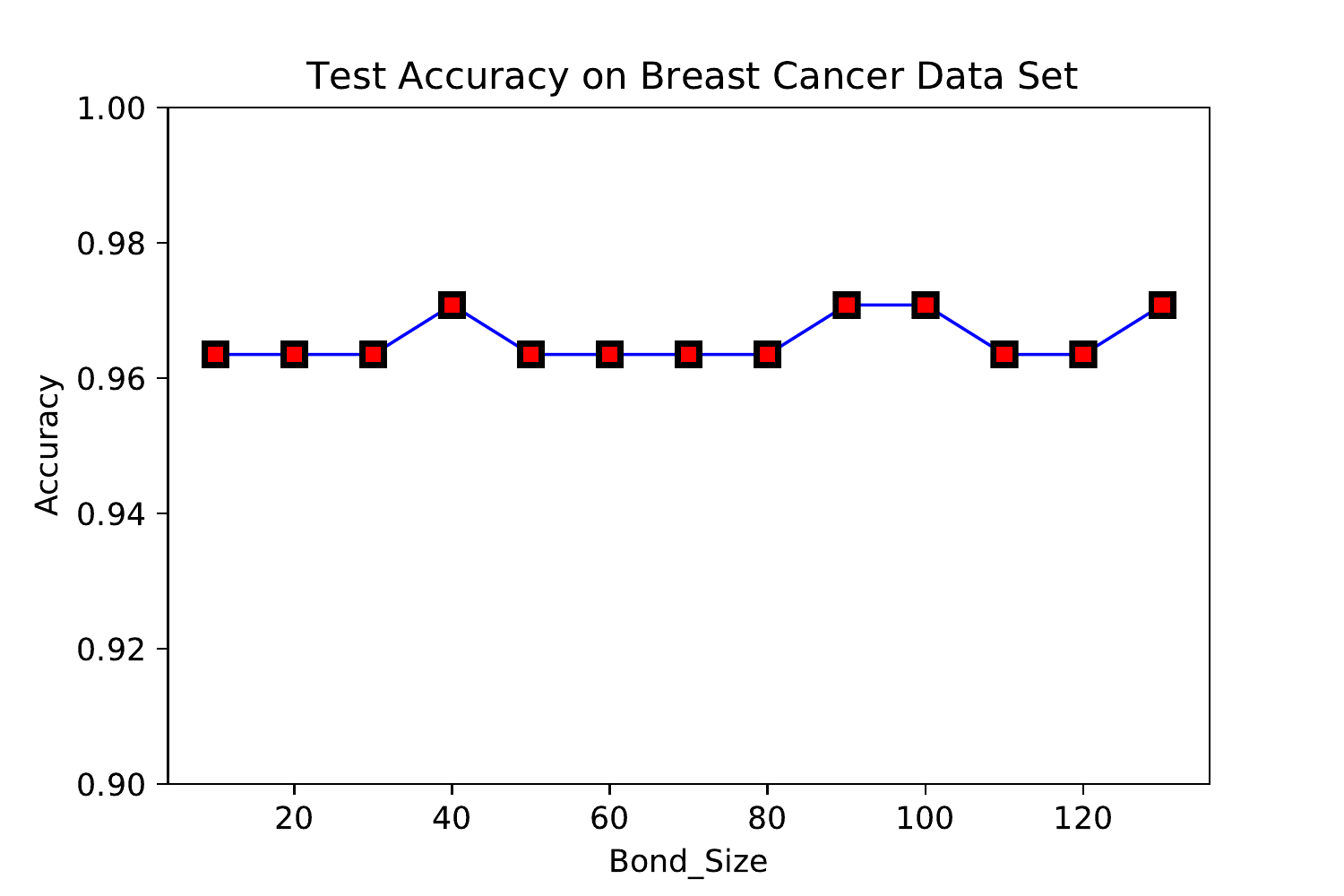}
  \caption{Breast Cancer Wisconsin data set}
  \label{fig: acc_breast_cancer_bond}
\end{subfigure} 
\caption{In (a), (b) and (c), we show the test accuracy of the Bayesian Tensor Network with different bond dimension on the MNIST, Phishing and Breast Cancer data set.}
\label{fig: mps_acc_bond}
\end{figure}

Bond dimension is a key hyperparameter in the MPS which controls the 'description' ability of the model \cite{collura2019descriptive}. In Fig. \ref{fig: mps_acc_bond}, we show the test accuracy of the Bayesian Tensor Network model with different bond dimension in different data set. 
We observe that as the bond dimension gets increased, the generalization ability of the model becomes better, namely the Bayesian Tensor Network model gets better prediction accuracy. 

\section{Conclusion}
We study the Bayesian framework of the Tensor Network and propose a robust initialization method. We use the toy, small and standard data set: Breast Cancer, Phishing website and MNIST data set to evaluate our initialization method and study the performance of the Bayesian Tensor Network model. We observe the Bayesian shrinkage in the parameters histogram plot and study the decision boundary of the Bayesian Tensor Network. We also explore the bond dimension in the Bayesian Tensor Network model. In practical application, we expect our model to take its own advantage in the small data set where overfitting problem can be solved by prior information introducing.

\section*{Acknowledgements}
The authors wish to thank David Helmbold, Hongyun Wang, Qi Gong, Torsten Ehrhardt and Francois Monard for their helpful discussions. 

\bibliographystyle{./ims.bst}
\bibliography{reference.bib}

\end{document}